\documentclass{article}

\usepackage{PRIMEarxiv}
\usepackage{kotex}
\usepackage[utf8]{inputenc}
\usepackage[T1]{fontenc}
\usepackage{hyperref}
\usepackage{url}
\usepackage{booktabs}
\usepackage{caption}
\usepackage{subcaption}
\usepackage{multirow}
\usepackage{amsmath}
\usepackage{array}
\usepackage{adjustbox}
\usepackage[sort&compress,numbers]{natbib}
\usepackage[ruled,vlined]{algorithm2e}
\usepackage{amsfonts}
\usepackage{nicefrac}
\usepackage{microtype}
\usepackage{lipsum}
\usepackage{fancyhdr}
\usepackage{graphicx}
\graphicspath{{media/}} 
\title{SCOPE: Stochastic and Counterbiased Option Placement for Evaluating Large Language Models}

\author{
  Wonjun Jeong, Dongseok Kim, Taegkeun Whangbo \\
  Department of Computer Engineering \\
  Gachon University \\
  Seongnam, South Korea\\
  \texttt{\{tp04045, jkds5920, tkwhangbo\}@gachon.ac.kr}
}

\begin{document}
\maketitle

\begin{abstract}
Large Language Models (LLMs) can achieve inflated scores on multiple-choice tasks by exploiting inherent biases in option positions or labels, rather than demonstrating genuine understanding. This study introduces SCOPE, an evaluation framework designed to measure and mitigate such selection bias in a dataset-independent manner. By repeatedly invoking a null prompt that lacks semantic content, SCOPE estimates each model's unique position-bias distribution. It then redistributes the answer slot according to the inverse-bias distribution, thereby equalizing the lucky-rate, the probability of selecting the correct answer by chance. Furthermore, it prevents semantically similar distractors from being placed adjacent to the answer, thereby blocking near-miss guesses based on superficial proximity cues. Across multiple benchmark experiments, SCOPE consistently outperformed existing debiasing methods in terms of stable performance improvements and showed clearer confidence distributions over correct options. This framework thus offers a new standard for enhancing the fairness and reliability of LLM evaluations.
\end{abstract}


\section{Introduction}
\subsection{Motivation: why multiple-choice bias matters}
Large Language Models (LLMs) have rapidly transformed the AI paradigm by surpassing human-level performance in diverse tasks such as document summarization, question answering, code generation, and logical reasoning \cite{achiam2023gpt, bubeck2023sparks, touvron2023llama, team2023gemini}. Recent state-of-the-art models—including ChatGPT, Claude, Gemini, and LLaMA—have evolved beyond simple statistical predictors to become tools capable of supporting real-world decision-making and creative tasks \cite{McKinseyGenAI2023, PwCOpenAI2024}.

However, these remarkable achievements often conceal underlying biases rooted in superficial cues \cite{mondorf2024beyond, du2024shortcut}. Recent studies, along with our preliminary experiments, reveal that LLMs frequently exhibit selection bias in multiple-choice (MCQ) settings: even without fully understanding the semantic content of the options, models tend to favor choices in certain positions or with shorter lengths \cite{zheng2023large, li2024anchored}. When the correct answer happens to be located at a preferred position, such bias can inflate accuracy, leading to an overestimation of the model’s actual language understanding ability \cite{li2024calibraeval}.

Reliable evaluation must begin by measuring and correcting such inherent selection bias through carefully controlled experimental design. However, most prior work has explored bias by modifying the dataset—shuffling the position of correct answers or replacing distractors—\cite{zheng2023large, wang2024look}, which captures only the model's interaction with the altered data rather than its intrinsic behavior. Consequently, any observed bias may stem more from the evaluation environment than from the model itself.

To address these limitations, this study proposes a novel approach that estimates a model's position bias distribution in a dataset-independent manner by injecting a large number of null prompts—inputs devoid of any semantic information—and recording the model’s positional preferences. The inverse-bias distribution is then used to sample answer slots, while the most semantically-similar distractor is placed probabilistically at the position farthest from the correct answer. This design minimizes reliance on shortcut strategies and enables an evaluation setting that isolates and tests the model’s genuine language understanding ability.

\subsection{Our contributions}
This study proposes a novel evaluation framework that structurally rearranges multiple-choice options by leveraging selection bias in Large Language Models (LLMs), without relying on any specific dataset. By injecting a large number of null prompts, we first extract each model’s intrinsic position bias distribution. The correct answer is then sampled from the corresponding inverse-bias distribution, and the most semantically-similar distractor is probabilistically relocated to a position far from the answer. As a result, the model is deprived of the opportunity to exploit superficial cues or shortcut strategies, and must instead rely solely on genuine language understanding to identify the correct answer.

Our contributions are summarized in four key aspects:
\begin{itemize}
\item \textbf{Dataset-independent position bias estimation:} We quantify each model’s internal position bias distribution through repeated trials with null prompts that contain no semantic information.
\item \textbf{Answer slot sampling from inverse-bias distribution:} By sampling the answer location from the inverse of the estimated bias distribution, we suppress the impact of lucky hits due to positional preference, enforcing a theoretical upper bound.
\item \textbf{Distance-weighted dispersion of semantically-similar distractors:} The most semantically-similar distractor is sampled via softmax to be placed far from the correct answer, thereby blocking near-miss guesses.
\item \textbf{Bias-resistant evaluation framework with theoretical guarantees:} The two modules together form SCOPE, a structural improvement over random shuffling or fixed layouts, with formal proofs demonstrating its ability to enable fair comparison across different LLMs.
\end{itemize}

\section{Related work}

\subsection{Benchmarks \& evaluation pipelines}
The evaluation of Large Language Models (LLMs) has advanced rapidly in the 2020s. Brown et al. \cite{brown2020language} demonstrated that a 175-billion-parameter model could solve a wide range of tasks using only simple prompts. Building upon this, Raffel et al. \cite{raffel2020exploring} unified all natural language problems into a text-to-text format and conducted large-scale experiments on 35 public datasets, thereby systematizing the generalizability of pretrained models. These two studies established three foundational axes for LLM benchmarks: multi-task coverage, large scale, and transferability.

Subsequent efforts focused on standardizing evaluation protocols and improving fairness. Hendrycks et al. \cite{hendrycks2020measuring} introduced the Massive Multitask Language Understanding (MMLU) benchmark, spanning 57 academic subjects to assess high-level expert knowledge. Talmor et al. \cite{talmor2019commonsenseqa} proposed CommonsenseQA (CSQA), a 9,500-question five-choice dataset derived from ConceptNet triples, requiring commonsense reasoning. Clark et al. \cite{clark2018think} released the AI2 Reasoning Challenge (ARC), comprising 7,787 multiple-choice science questions from elementary and middle school exams, and showed that existing QA models perform near chance level. Mihaylov et al. \cite{mihaylov2018can} introduced OpenBookQA, which combines 1,329 scientific facts with 5,957 four-choice questions that demand multi-hop reasoning using open-book knowledge. Srivastava et al. \cite{srivastava2023beyond} presented BIG-bench, a meta-benchmark containing 204 diverse tasks designed to capture both task difficulty spectrum and the performance–model-size trend. Liang et al. \cite{liang2022holistic} developed HELM, which evaluates models across seven axes—response quality, bias, readability, inference length, and more—with open logs to greatly enhance transparency in data, models, and metrics.

Since 2023, there has been increasing emphasis on diversity in domain, language, and modality, as well as alignment with real user preferences. Zhang et al. \cite{zhang2023m3exam} constructed a human-level benchmark across nine languages, four modalities, and twelfth-grade exam questions. Chen et al. \cite{zhong2024agieval} examined model performance in high-stakes domains, including national licensing exams for lawyers, doctors, and teachers in China and the United States. Pal et al. \cite{pal2022medmcqa} introduced a massive medical benchmark with 194,000 questions spanning 21 medical subjects from India’s AIIMS and NEET-PG exams. Welbl et al. \cite{welbl2017crowdsourcing} curated 13.7K high-quality MCQs for K–12 science via crowdsourcing, demonstrating indistinguishability from actual test questions. Angelopoulos et al. \cite{li2024live} processed real Chatbot Arena conversations to build a balanced dynamic evaluation set, while Chiang et al. \cite{chiang2024chatbot} converted over 300,000 pairwise preference votes into Elo scores to establish a new metric for real-time, human-preference-based ranking. Wang et al. \cite{wang2024large} showed that model rankings can be arbitrarily altered depending on response order and proposed MEC, BPC, and HITLC metrics to better align with human judgments. Finally, Wu et al. \cite{wu2025apbench} introduced domain-specific evaluations in orbital mechanics and navigation engineering, conducting the first systematic assessment of LLMs’ numerical accuracy and adherence to physical constraints in science and engineering.

\subsection{Position \& label bias in MCQs}

The tendency of LLMs to prefer certain positions regardless of answer content—referred to as selection bias—was first measured at scale by Zheng et al. \cite{zheng2023large}. Saito et al. \cite{saito2025answer} further demonstrated the generality of this issue by identifying similar vulnerabilities in document-based knowledge extraction tasks. Yang et al. \cite{yang2025option} revealed that the option label itself can be a source of error, while Pezeshkpour and Hruschka \cite{pezeshkpour2024large} showed that simply shuffling answer order can boost GPT-4 accuracy by up to 75\%, with multi-evidence calibration (MEC) and majority voting strategies recovering performance by up to 8 percentage points—reaffirming the severity of position bias in LLMs. Zheng et al. \cite{zheng2309large} proposed PriDe, a method that estimates and separates option ID priors using only 5\% of the data, offering a much cheaper alternative to full permutation for bias removal. Li et al. \cite{li2024calibraeval} introduced CalibraEVAL, a correction function that preserves order without labels during inference, reducing variance and improving accuracy across a wide range of LLMs and benchmarks.

Zong et al. \cite{zong2024fool} showed that randomly shuffling option order alone can cause state-of-the-art LLMs to degrade to near-random guessing levels. In contrast, Liusie et al. \cite{liusie2024teacher} proposed a computationally efficient correction method based on teacher–student knowledge distillation to learn permutation-invariance. Structural and interpretable approaches have also been explored: Choi et al. \cite{choi2024mitigating} removed internal nodes responsible for bias and injected auxiliary options to correct internal representations. Wang et al. \cite{wang2024look} found that evaluating based on generated text rather than logits makes models less sensitive to positional bias, while Li and Gao \cite{li2024anchored} conducted a mechanical analysis and mitigation of the “anchored answer” bias in the GPT-2 series. Finally, Wei et al. \cite{wei2024unveiling} proposed a unified metric that captures both positional and token-level biases, laying the groundwork for a standardized evaluation across future studies.

\subsection{Prompt-time reasoning \& inference-time debiasing}

Efforts to induce genuine reasoning in LLMs—rather than reliance on superficial biases—gained traction with Wei et al.’s “Let’s think step by step” approach \cite{wei2022chain}, which led to substantial improvements in mathematical, commonsense, and logical reasoning tasks through a single prompt phrase. Kojima et al. \cite{kojima2022large} further demonstrated that such gains could be achieved even without few-shot demonstrations, showing that prompting alone can elicit complex reasoning.

However, a single reasoning path can easily become trapped in error. To address this, Wang et al. \cite{wang2022self} introduced reasoning ensembles, in which multiple Chain-of-Thought (CoT) trajectories are sampled and aggregated via majority voting, enhancing logical consistency. Zelikman et al. \cite{zelikman2024star} proposed a self-supervised refinement strategy in which the model filters high-quality chains from its own generations and iteratively retrains on them, thereby reinforcing reliable reasoning.

Another line of work reduces bias by decomposing problems into sub-problems. Zhou et al. \cite{zhou2022least} advanced a least-to-most prompting strategy, where models solve simpler sub-questions before tackling more complex reasoning steps, improving performance in arithmetic and logic. Yao et al. \cite{yao2023react} proposed ReAct, which interleaves reasoning and acting, allowing models to update incorrect assumptions in real-time based on environmental feedback. Li et al. \cite{li2024debiasing} identified semantic ambiguity as a key source of demonstration and order bias, and proposed a hybrid approach that combines ambiguity-score-based example reordering with self-explaining prompts to enhance both robustness and accuracy across six datasets. Structured techniques involving pseudocode have also emerged: Chen et al. \cite{chen2022program} translated natural language reasoning into executable Python code, and Beurer-Kellner et al. \cite{beurer2024prompt} proposed condensing lengthy chains into compact pseudocode sketches for more efficient problem solving.

Several approaches also aim to reduce both bias and harmful outputs by enforcing normative or logical constraints. Madaan et al. \cite{madaan2023self} automated a generate–review–revise loop to enable self-correction. Bai et al. \cite{bai2022constitutional} implemented Constitutional AI, appending a set of normative rules to the prompt to retroactively censor bias and toxic content. Lu et al. \cite{lu2021neurologic} applied logic constraints dynamically during beam search to suppress contradictions and overgeneralized reasoning.

\subsection{Probability calibration \& lucky-hit mitigation}

Beyond accuracy, the confidence of a model’s predictions has long been recognized as critical for ensuring safety and fairness. Desai et al. \cite{desai2020calibration} reported that BERT-based models often exhibit over-confidence, assigning probabilities that exceed actual correctness. Zhao et al. \cite{zhao2021calibrate} estimated fixed model biases using information-less N/A prompts and applied a calibration vector, which stabilized GPT-3’s few-shot performance by an average of 5 percentage points.

Subsequent work introduced finer-grained calibration methods. Xie et al. \cite{xie2024calibrating} proposed Adaptive Temperature Scaling (ATS), which adjusts the temperature parameter per token to restore distorted probability distributions following Reinforcement Learning with Human Feedback (RLHF). Lyu et al. \cite{lyu2025calibrating} addressed the lucky hit problem by sampling multiple responses for the same question and converting answer consistency into probability estimates, reducing Expected Calibration Error (ECE) by 40

Benchmarks have also emerged to compare calibration methods more broadly. Vashurin et al. \cite{vashurin2025benchmarking} introduced the UQ benchmark, which consolidates 11 tasks and re-evaluates 12 calibration methods under identical conditions. Shen et al. \cite{shen2024thermometer} proposed a universal calibration approach that generalizes across new tasks using multi-task and multimodal data.

Parallel research has focused on eliminating luck-based variance to improve confidence reliability. Cecere et al. \cite{cecere2025monte} randomized the temperature parameter to promote sampling diversity and achieved stable uncertainty estimates without hyperparameter tuning. Zhu et al. \cite{zhu2025charm} leveraged Arena Elo scores to debias reward models, reducing human preference estimation errors by over 20\%. Lastly, Xiao et al. \cite{xiao2025restoring} incorporated ECE-based normalization directly into the loss function, allowing over-confident models trained with RLHF to recover reliability without sacrificing performance. From an applied perspective, Mora-Cross and Calderon-Ramirez \cite{mora2024uncertainty} demonstrated that combining Monte Carlo Dropout with ECE calibration can effectively mitigate risk in real-time decision support settings.

\subsection{Cognitive-inspired evaluation \& memory stress tests}
\label{subsec:cognitive_bias}

There is growing interest in evaluating whether LLMs exhibit cognitive capabilities similar to those of humans. Kosinski et al. \cite{kosinski2023theory} showed that GPT-3.5 and GPT-4 can solve classical false-belief tasks, suggesting that Theory of Mind (ToM) may emerge abruptly with increased parameter scale. In contrast, Sap et al. \cite{sap2022neural} argued that ToM performance is highly dependent on dataset design, citing low scores on SOCIALIQA \cite{sap2019social} and similar benchmarks. Meanwhile, Gendron et al. \cite{gendron2024large} experimentally demonstrated that LLMs struggle to fully comprehend or accurately reason through abstract logic puzzles, and that cognitive tasks rooted in human psychology effectively expose these limitations.

Efforts have also been made to trace models’ cognitive development. Wang et al. \cite{wang2025coglm} evaluated 14 models across 1,220 Piagetian-stage tasks and found that model size and training objectives influence developmental trajectories. Lin et al. \cite{lin2025valphasocial} assessed the explanatory social reasoning of vision-language models by prompting them to “explain their thinking” when interpreting image-text scenes.

Other studies have explored how LLMs perform actions in external environments. Liu et al. \cite{deng2023mind2web} introduced Mind2Web, measuring task completion rates over 2,350 real-world web page tasks involving instructions, DOM manipulation, and feedback. A follow-up study \cite{gou2025mind2web} added long-horizon navigation and automatic grading to standardize comparisons.

Memory stress tests targeting long-context capabilities have become increasingly sophisticated. Tay et al. \cite{tay2020long} found that standard Transformers degrade significantly beyond 8k tokens when handling sequences of text, images, and formulas up to 16k tokens. Rühle et al. \cite{xia2025minerva} proposed PMT to separately measure serial and spatial memory, while Zhao et al. \cite{an2024eval} conducted fine-grained analyses of performance breakdown points by prompting models to summarize, answer questions, and complete code over 200k-token documents. Nelson et al. \cite{nelson2024needle} showed that recovering hidden strings embedded in contexts of hundreds of thousands of tokens requires external memory integration, underscoring the need for true long-term memory mechanisms.

Finally, Chan et al. \cite{chan2024conversational} simulated a witness-interrogation scenario and found that LLM-based chatbots significantly reinforced false suggestions, tripling the probability of false memory reporting. The study underscores the need for safeguards when deploying LLMs in high-stakes domains.

\section{Preliminary experiments}
The ultimate objective of this study is to construct an evaluation framework that controls selection bias in LLMs and prevents models from arriving at correct answers through shortcut strategies. Before formally introducing our approach, we conducted a preliminary experiment using a 2 × 2 factorial design to evaluate how effective basic debiasing methods—such as removing label and position cues—are in practice.

The two experimental factors were the presence of answer labels (L) and whether the order of options was fixed (F). The negation operator $\lnot$ indicates the removal or reversal of a given factor. Accordingly, the four experimental conditions were as follows: L + F (Baseline), $\lnot$L + F (label removed), L + $\lnot$F (order randomized), and $\lnot$L + $\lnot$F (fully randomized).

We applied these conditions to Claude 3-haiku using the MMLU benchmark. Both $\lnot$L + F and L + $\lnot$F partially mitigated selection bias. Table~\ref{tab:performance_comparison} summarizes the changes in KL divergence (KLD) and accuracy across the four conditions. In the Baseline condition, KLD was 0.0191, which dropped to 0.0101 under $\lnot$L + F and to 0.0033 under L + $\lnot$F. In the fully randomized condition ($\lnot$L + $\lnot$F), KLD further decreased to 0.0018, indicating that the model's selection rate distribution converged most closely to the ideal uniform distribution.

Nevertheless, the fact that KLD did not reach zero suggests that deeper response patterns—likely formed during pretraining or induced by prompt structure—remain active even after surface-level cues are removed. Meanwhile, accuracy dropped dramatically from 0.676 (Baseline) to 0.238 under the \textbf{$\lnot$L + $\lnot$F} condition. This confirms that when label and order cues are eliminated, the model struggles to solve problems and can no longer rely on shortcut strategies.

These findings empirically demonstrate that naive randomization alone is insufficient to fully eliminate bias and highlight the necessity for a more sophisticated mechanism for bias control. Full experimental results are presented in Appendix~\ref{app:preliminary_condition}.

\begin{table}[ht!]
\centering
\caption{Impact of label hiding \& option shuffling on MMLU (selection rate, KLD, accuracy)}
\label{tab:performance_comparison}
\begin{tabular}{ccccc}
\toprule
Metric & Baseline & $\lnot$L + F & L + $\lnot$F & $\lnot$L + $\lnot$F \\
\midrule
Selection Rate & \raisebox{-0.5\height}{\includegraphics[width=0.15\textwidth]{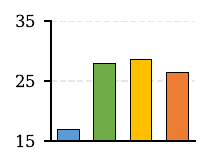}} & \raisebox{-0.5\height}{\includegraphics[width=0.15\textwidth]{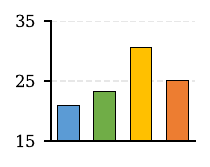}} & \raisebox{-0.5\height}{\includegraphics[width=0.15\textwidth]{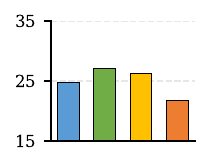}} & \raisebox{-0.5\height}{\includegraphics[width=0.15\textwidth]{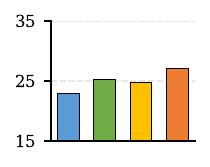}}
\\
KLD   & 0.0191 & 0.0101 & 0.0033 & 0.0018 \\
Accuracy & 0.6760 & 0.6710 & 0.6460 & 0.2380 \\
\bottomrule
\end{tabular}
\end{table}

We also observed that randomly redistributing the answer position reveals another critical issue. When evaluating Claude 3-haiku on the MMLU dataset, we found that if the correct answer is incidentally placed in the position most frequently selected by the model, it can be answered correctly not through semantic understanding, but by relying on the model's positional preference. This leads to an artificial inflation of accuracy due to the lucky-hit effect.

To quantify how much this lucky hit phenomenon distorts performance, we conducted an additional experiment under the Low-Bias Placement (LBP) condition, in which correct answers were deliberately placed only in positions where the model’s selection bias was empirically measured to be low. Under the Baseline setting, accuracy was 0.6700, but dropped significantly to 0.5330 under the LBP condition—a decrease of more than 0.13 points.

This result clearly demonstrates that answer position alone can lead to an overestimation of model performance. Therefore, to ensure valid evaluation of actual model capability, it is essential to impose placement constraints that account for position bias. Full experimental results are provided in Appendix~\ref{app:preliminary_lbp}.

\begin{table}[ht!]
\centering
\caption{Accuracy drop after re-locating answers to least-preferred slots (LBP)}
\label{tab:lbp_comparison}
\begin{tabular}{ccc}
\toprule
Metric & Baseline & LBP \\
\midrule
Selection Rate & \raisebox{-0.5\height}{\includegraphics[width=0.15\textwidth]{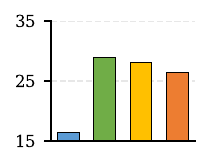}} & \raisebox{-0.5\height}{\includegraphics[width=0.15\textwidth]{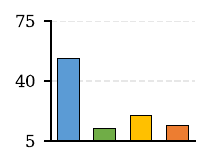}} \\
Accuracy & 0.6700 & 0.5330 \\
\bottomrule
\end{tabular}
\end{table}

Additionally, we examined the effect of different placement strategies for the most semantically-similar distractor (SSD) relative to the correct answer. As shown in the results for Claude 3-haiku (Table~\ref{tab:claude_ssd_accuracy}), the SSD selection rate under the unconstrained Baseline condition was 0.422. However, when the SSD was explicitly placed adjacent to the correct answer (Condition A), the rate increased to 0.436—an increase of 1.4 percentage points—indicating a higher tendency to confuse the SSD with the correct answer.

In contrast, under the $\lnot$A condition, where the SSD was forcibly placed away from the correct answer, the SSD selection rate dropped sharply to 0.334, an 8.8 percentage point decrease. This suggests that such misinterpretations were substantially mitigated. Notably, accuracy changed by no more than ±0.8 percentage points across conditions, demonstrating that the positional constraint suppresses near-miss guesses without impairing the model's core reasoning ability.

These findings provide indirect evidence that Claude employs heuristics based on both semantic similarity and relative position among options. They also imply that even simple positional constraints—such as those implemented in the SS module—can effectively and precisely regulate the strength of bias patterns. Full results of the SSD adjacency placement experiment are presented in Appendix~\ref{app:preliminary_ssd}.

\begin{table}[ht!]
\centering
\caption{Effect of forcing semantically similar distractors (SSD) adjacent/non-adjacent}
\label{tab:claude_ssd_accuracy}
\begin{tabular}{ccccc}
\toprule
Model & Metric & Baseline & A & $\lnot$A \\
\midrule
\multirow{2}{*}{Claude (3-haiku)} & SSD selection rate & 0.422 & 0.436 & 0.334 \\
                                  & Accuracy     & 0.678 & 0.670 & 0.671 \\
\bottomrule
\end{tabular}
\end{table}

Through these preliminary experiments, we confirmed that simple interventions—such as removing answer labels, randomizing option order, or redistributing the correct answer’s position—are insufficient to suppress LLMs’ shortcut strategies or eliminate selection bias. This underscores the need for a new evaluation design that independently measures the model’s positional preferences outside the dataset, and uses this information to systematically reassign the positions of both correct answers and distractors.

The following section introduces our proposed evaluation framework, which is designed with this motivation in mind.

\section{Methodology}
To properly measure and control selection bias in LLMs, it is essential to disentangle the model’s intrinsic behavioral tendencies from the structure of the dataset itself. Prior studies have typically inferred bias by shuffling answer positions or modifying distractor compositions, but these approaches are limited in that they observe model responses only after semantically meaningful questions and choices have already been provided. When the measurement instrument is itself biased, the results are inevitably distorted.

This paper introduces SCOPE, a two-module evaluation framework consisting of Inverse-Positioning (IP) and Semantic-Spread (SS).

The IP module estimates the model’s position preferences independently by issuing a large number of null prompts—inputs that contain no semantic content. For example, a prompt like “You must choose one. If you had to pick, which would it be?” is followed by randomly generated alphabetical options, thereby ensuring that no meaningful judgment can be made. The frequency with which each position is selected is treated as the model’s empirical position bias distribution, and the correct answer’s location is then sampled inversely proportional to this distribution. As a result, models are discouraged from relying on preferred positions to make lucky guesses and are instead forced to depend on actual language understanding and reasoning.

The SS module suppresses near-miss guesses by ensuring that the semantically similar distractor (SSD) is placed far from the correct answer. All options are projected into a semantic space using Sentence-BERT embeddings \cite{reimers2019sentence}, and the SSD is identified via cosine similarity \cite{schutze2008introduction}. The SSD’s position is then sampled from a distribution that increases in probability with distance from the correct answer. This reduces the likelihood that the SSD appears adjacent to the correct option, thereby minimizing the risk that the model selects it based solely on semantic proximity.

\begin{figure}[ht!]
    \centering
    \includegraphics[width=\textwidth]{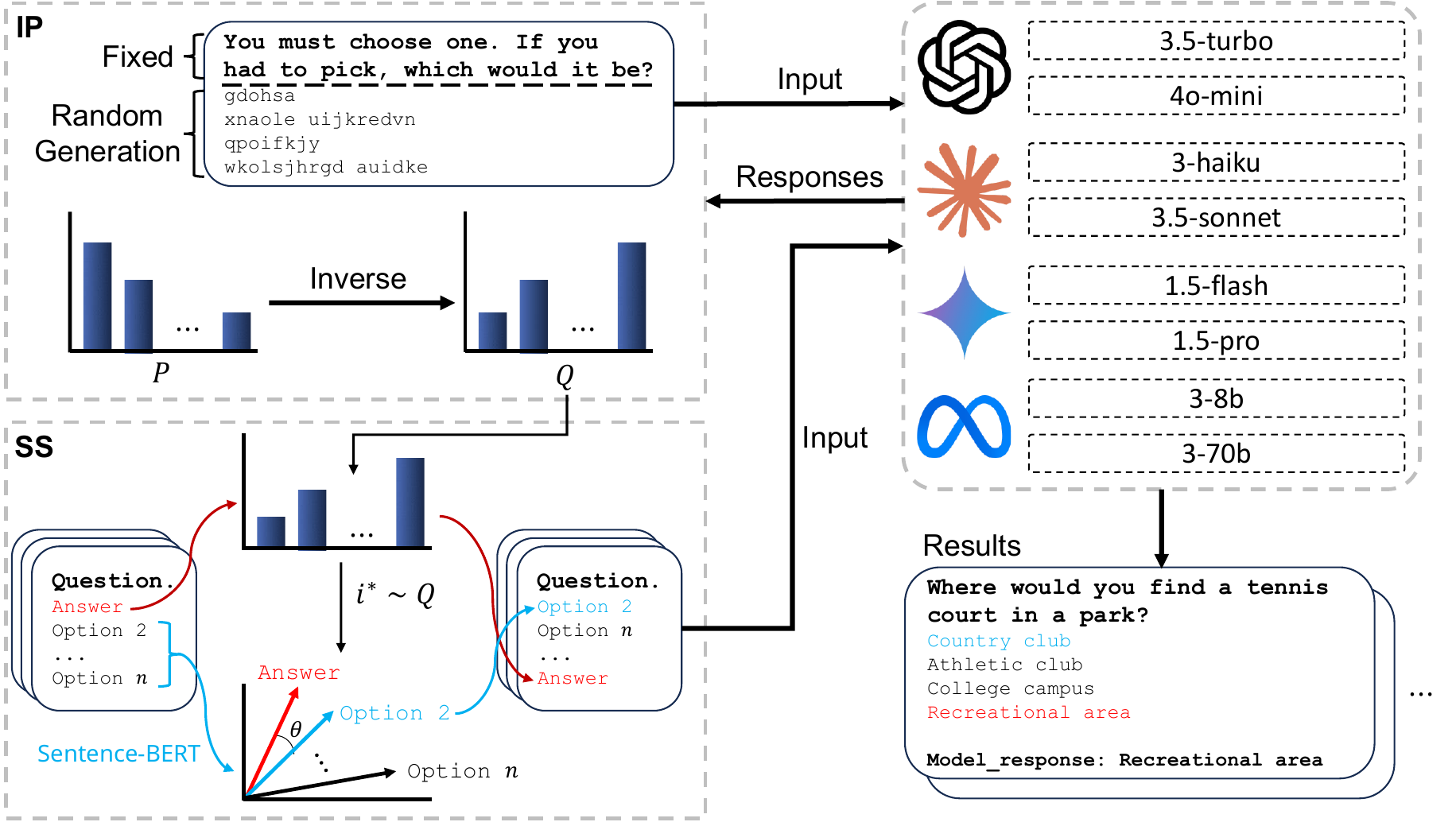}
    \caption{Overall architecture of the SCOPE framework (IP + SS)}
    \label{fig:framework}
\end{figure}

By equalizing the lucky-rate arising from position bias and blocking distractor selections driven by shallow semantic cues, SCOPE reveals a model’s true language understanding ability more fairly than simple random shuffling or label removal. Furthermore, prior to evaluation we eliminate all option labels—or, equivalently, replace them with an identical placeholder such as an en-dash—thereby neutralizing any residual bias that label cues might introduce. The overall architecture of SCOPE is illustrated in Figure~\ref{fig:framework}, the pseudocode is provided in Algorithm~\ref{alg:scope}, and detailed descriptions of the IP and SS modules are given in Sections~\ref{subsec:IP_module} and~\ref{subsec:SS_module}, respectively.

\begin{algorithm}[ht!]
\caption{SCOPE pipeline}
\label{alg:scope}

\SetKwInput{KwIn}{Input}
\SetKwInput{KwOut}{Output}

\KwIn{LLM $f$; dataset $\mathcal{D}$ ($N$ items, $n$ options);
      Null-prompt budget $M$;
      Neutral-prompt template $u$;
      Encoder $\phi$}
\KwOut{Predictions $\hat{\mathbf y}$ and Evaluation metrics $\mathcal{E}$}

$\;Q \leftarrow \mathbf{IP}\bigl(f,\,M,\,u\bigr)$\tcp*[r]{inverse-bias dist.}

\For{$(x,\,\mathcal{O},\,a)\ \in\ \mathcal{D}$}{      
    $\bigl(i^\star,\,j_{\text{SSD}}\bigr) \leftarrow
        \mathbf{SS}\bigl(\mathcal{O},\,a,\,Q,\,\phi\bigr)$\tcp*[r]{distance-based SSD}
    $\tilde{\mathcal{O}} \leftarrow
        \Pi\bigl(\mathcal{O},\,i^\star,\,j_{\text{SSD}}\bigr)$\;
    $\hat{y} \leftarrow f\!\bigl(x,\,\tilde{\mathcal{O}}\bigr)$\;
    $\Delta\bigl(\hat{y}\bigr)$\;
}
\KwRet{$\hat{\mathbf y}$, $\mathcal{E}$}\;
\end{algorithm}

\subsection{Inverse-Positioning (IP)}
\label{subsec:IP_module}
Large Language Models (LLMs) exhibit a position bias in multiple-choice questions, whereby they disproportionately favor certain answer slots regardless of their semantic content. Both previous research and our preliminary experiments show that this bias persists even when the order of the options is shuffled or the answer label is removed. Consequently, if correct answers are accidentally concentrated in the preferred positions within the evaluation set, the model may achieve high accuracy without properly understanding the question, leading to an overestimation of its performance. The IP module of SCOPE addresses this risk by reassigning correct answers less frequently to highly biased positions.

To begin, the model is prompted thousands of times with a null prompt, which contains no semantic content, to estimate the selection rate of each answer slot:

\[
P = (p_1, \dots, p_n), \qquad \sum_{i=1}^{n} p_i = 1
\]

Then, the answer slot is determined based on probabilities proportional to the inverse of $P$. For each position $i$, we define:

\[
q_i = \frac{1}{\sum_{j=1}^{n} 1/p_j}, \qquad Q = (q_1, \dots ,q_n)
\]

and the answer slot $i^*$ is sampled as $i^* \sim Q$ for each question. Because $q_i$ becomes larger for less-preferred positions, correct answers are more likely to be placed in answer slots with smaller position bias. This procedure yields two effects. First, it effectively eliminates the lucky hit by bounding the lucky-rate $\ell$—i.e., the probability of guessing the correct answer based solely on positional cues—to at most $1/n$. Second, since the answer slot is resampled independently for each question, the global concentration of correct answers in specific positions across the evaluation set disappears. As a result, to achieve high performance, the model must interpret and reason about the meaning of the choices, and SCOPE accurately measures only the model's true understanding after neutralizing selection bias.

The IP module requires only a one-time measurement of position bias and is simple to implement. Moreover, because it reflects each model's individual position bias, it can be easily extended to construct evaluation sets tailored to specific models.

\subsection{Semantic-Spread (SS)}
\label{subsec:SS_module}
While the inverse-bias positioning reduces the likelihood of correct answers being concentrated in preferred slots, another source of evaluation distortion remains. If the semantically similar distractor (SSD) is placed directly adjacent to the correct answer, the model may engage in near-miss guessing by relying solely on the proximity of the two options. In such cases, the model may achieve high accuracy without truly understanding the content, leading to an overestimation of its language comprehension.

To address this issue, the Semantic-Spread (SS) module in SCOPE assigns SSDs to positions that are physically distant from the correct answer, with probability increasing in proportion to their separation. The procedure is as follows. First, we obtain sentence embeddings for all options and compute cosine similarity between the correct answer and each distractor. The distractor with the highest similarity score is designated as the SSD.

Once the correct answer has been placed at position $i^*$ through inverse-bias sampling, we define a candidate set for SSD placement: $S = \left\{ 1, \dots , n\right\} \setminus \left\{ i^* \right\}$. For each candidate position $j \in S$, we compute its absolute distance from the answer position $d = \left | i^* - j \right |$ and assign an exponential weight:

\[
w_j = \exp(d_j)
\]

Normalizing these weights gives the final distribution:

\[
r_j = \frac{w_j}{\sum_{k \in S} w_k}, \qquad j \in S
\]

from which the SSD position is sampled. Because the probability increases exponentially with distance, SSDs are on average placed farther from the correct answer.

If multiple distractors exhibit high semantic similarity, only the most similar one is designated as the SSD and treated with this sampling process. The remaining distractors are assigned to the leftover positions via uniform random placement.

This exponentially weighted dispersion systematically increases the expected distance between correct answers and semantically similar distractors, preventing clusters of closely related choices from forming around the answer. As a result, the model’s ability to exploit shortcut strategies is diminished, and the SCOPE framework more accurately reveals the model’s true capacity for semantic understanding.

\section{Theoretical analysis}
This section demonstrates how SCOPE mathematically cancels out the model's position bias and interprets the extent to which the dispersion of semantically-similar distractors suppresses shortcut guessing. We first define the notation, then present Theorem 1 to demonstrate the unbiasedness of answer slot sampling, and Proposition 1 to show the effect of distractor dispersion. The main proofs are included in the main text, with detailed derivations provided in Appendices~\ref{app:proof_bias_cancel} and~\ref{app:proof_SSD}.

\subsection{Notation \& assumptions}
\begin{itemize}
    \item $n$: the number of choices presented in a single question
    \item $P = [p_1, \dots, p_n]$: position bias distribution measured using a null prompt $(\sum_{i} p_i = 1, \quad p_i > 0)$
    \item $Q = [q_1, \dots, q_n]$: inverse-bias distribution $(q_i = \frac{1/p_i}{\sum_{j}1/p_j})$
    \item $i^*$: answer slot random variable $(i^* \sim Q)$
    \item $\ell$: the lucky-rate—the probability that the model selects the correct answer based solely on position bias without interpreting the content $(\ell = \sum_{i=1}^{n} p_i q_i)$
\end{itemize}

\subsection{Position-bias cancellation theorem}
\subsubsection*{Theorem.}
\[
\ell = \frac{1}{n}
\]

\subsubsection*{Proof.}
\[
\ell = \frac{\sum_{i=1}^{n} p_i \cdot \frac{1}{p_i}}{\sum_{j=1}^{n}\frac{1}{p_j}} = \frac{n}{\sum_{j=1}^{n}\frac{1}{p_j}}
\]

By the inequality between harmonic mean and arithmetic mean (HM $\leq$ AM), we have $\sum_{j}\frac{1}{p_j} \geq n^2$, which implies

\[
\ell \leq \frac{n}{n^2} = \frac{1}{n}
\]

Equality holds if and only if $p_1 = \dots = p_n = 1/n$, that is, when the model has no position bias.

\subsubsection*{Corollary.}
If the observed accuracy under SCOPE exceeds $1/n$, the excess is guaranteed to result from the model’s understanding of the content. Moreover, since the lower bound of $\ell$ is $p_{\min} = {\min}_i p_i$, the single bound $p_{\min} \leq \ell \leq \frac{1}{n}$ holds.

\subsection{Distance-weighted distractor dispersion}
Let the answer slot be $i^*$, and let the distance from each distractor position $j$ be defined as $d_j = \left | i^* - j \right |$. SCOPE defines an exponentially weighted distribution using $w_j = \exp(d_j)$ to place the semantically-similar distractor away from the answer:

\[
r_j = \frac{w_j}{\sum_{k \neq i^*} w_k}
\]

\subsubsection*{Proposition.}
\begin{center}
For $n \geq 3$, \quad it holds that $\mathbb{E}_R \left[ d \right] > \mathbb{E}_{\text{unif}} \left[ d \right]$.
\end{center}

\subsubsection*{Proof.}
Since $\exp(d)$ is monotonically increasing and convex in $d$, Jensen's inequality implies that the exponential-weighted distribution $R$ shifts probability mass toward larger distances. Therefore, the expected distance $\mathbb{E}_R \left[ d \right]$ is always greater than that under uniform placement.

\subsubsection*{Corollary.}
When semantically-similar distractors are concentrated near the correct answer, the model may exploit a shortcut strategy to guess correctly. The exponential dispersion structurally blocks this pattern and reduces the likelihood of near-miss guesses.

\subsection{Key findings \& outlook}
Theorem 1 guarantees that inverse-bias sampling limits the effect of position bias to a lucky-rate no greater than $1/n$. Proposition 1 shows that exponential distance-based dispersion spreads distractors farther from the correct answer, thereby increasing evaluation difficulty. Together, these results mathematically support the design philosophy of SCOPE: to neutralize innate position bias and measure only true content understanding. With this theoretical foundation established, the next section empirically validates these effects through large-scale experiments and an ablation study.

\section{Experimental design}
This study systematically removes selection bias exhibited by LLMs in multiple-choice question answering and evaluates, through repeated trials, whether the model truly understands the question—i.e., whether it solves the problem without guessing. The core assumption is as follows: once bias-related cues are eliminated, the model can no longer rely on probabilistically favorable positions to make random selections. Thus, for questions it knows, the model will consistently select the correct answer across repeated queries, whereas for questions it does not know, it will persistently choose the same incorrect answer. By verifying such consistent behavioral patterns, the study aims to empirically demonstrate that the model’s linguistic competence can be clearly distinguished from its areas of misunderstanding.

\subsection{Collecting response patterns via repetition}
All bias-mitigation methods follow a common input pipeline. The goal is to observe how consistently a model responds to a single question. Each item in the dataset is first transformed into a method-specific prompt, tailored to reflect the corresponding bias-mitigation technique. During this transformation, each method applies its own correction logic for selection bias, while preserving the original structure of the question. The resulting prompt is then fed into the language model five times consecutively without modification. The five responses are compared to measure the model's selection preference and response consistency. The number of repetitions is fixed at five, which is the statistically minimal value ensuring meaningful results under the constraints of computational cost and time.

\subsection{Comparison methods}
This study compares a total of seven conditions under a unified experimental protocol. First, the \textbf{Baseline} condition refers to the unmodified evaluation set without any bias correction. \textbf{SCOPE} denotes our proposed method for eliminating selection bias. In addition, we reproduce five representative methods from recent literature for comparison:

\begin{itemize}
    \item \textbf{CalibraEval (CalibEV, \cite{li2024calibraeval})}: Reformulates selection bias as a calibration problem over the predicted probability distribution and adjusts it using a non-parametric unsupervised method. Following the original setting, we apply the correction function after three repeated estimations.
    
    \item \textbf{Debiasing In-Context Learning (DI, \cite{li2024debiasing})}: Injects explicit instructions into the prompt that discourage selection bias, encouraging the LLM to self-correct. We use the exact instruction format specified in the original paper.
    
    \item \textbf{Evidence Calibration (EC, \cite{wang2024large})}: Requires the model to first generate reasoning before scoring candidate responses, thereby reducing inconsistencies between reasoning and final predictions. We follow the two-step API call procedure and use $k=3$ reasoning samples.
    
    \item \textbf{Majority Voting (MV, \cite{pezeshkpour2024large})}: A simple ensemble approach that presents the same question with multiple permutations and selects the most frequent answer as the final prediction. In our implementation, we randomly shuffle the answer options 10 times.
    
    \item \textbf{Prior Debiasing (PriDe, \cite{zheng2309large})}: Estimates a prior over option identifiers from a small subset of examples and corrects for bias across the entire evaluation set. As in the original paper, we reserve 5\% of the test set for prior estimation.
\end{itemize}

All methods were evaluated using the same language model, the same evaluation set, and identical decoding parameters (temperature 1, greedy decoding). Post-processing steps and hyperparameters follow the default settings recommended in their respective original implementations.

\subsection{Response-pattern taxonomy}
Each question is categorized into one of four types based on the model’s five repeated responses. If the same option is chosen at least three times, we consider it a case of preference; if the same option is chosen all five times, we interpret it as consistency. The four resulting empirical metrics—Pr-T, Pr-F, Co-T, and Co-F—correspond to “preferred correct answer,” “preferred distractor,” “consistent correct answer,” and “consistent distractor,” respectively. Among these, Co-T and Co-F serve as key indicators of whether the LLM truly understands the content. A consistent correct response (Co-T) suggests that the model possesses stable, learned knowledge. In contrast, a consistently incorrect response (Co-F) implies that the model has an incorrectly formed internal concept about the topic of the question. Questions falling within the Pr category exhibit residual randomness, indicating that the model selects an answer without full certainty.

\begin{table}[ht!]
\centering
\caption{Preference (Pr) \& Consistency (Co) Metrics — definitions of Pr-T, Pr-F, Co-T, Co-F}
\label{tab:preference_consistency}
\begin{tabular}{cccc}
\toprule
\multicolumn{2}{c}{Metric} & Definition & Interpretation \\
\midrule
\multirow{2}{*}{Preference (Pr)} 
  & T & Chose the correct answer at least 3 out of 5 times & Preferred correct answer \\
  & F & Chose the correct answer at most 2 out of 5 times  & Preferred distractor \\
\multirow{2}{*}{Consistency (Co)} 
  & T & Chose the correct answer in all 5 trials            & Consistent correct answer \\
  & F & Chose a distractor in all 5 trials                  & Consistent distractor \\
\bottomrule
\end{tabular}
\end{table}

\subsection{Metrics derived from repeated trials}
Simple accuracy alone cannot distinguish whether a correct answer from an LLM was the result of true understanding or mere guessing. To address this, we reformulate standard classification metrics (Precision, Recall, F1) to define two metric families. The \textbf{answer-metric family} captures the degree to which the model exhibits certainty toward the correct answer, while the \textbf{distractor-metric family} captures its certainty toward incorrect answers.

Answer Precision (AP) is defined as Co-T / (Co-T + Co-F), representing the proportion of consistent responses that are correct. Answer Recall (AR) is defined as Co-T / (Pr-T + Co-T), measuring how often a preference for the correct answer results in full certainty. Their harmonic mean yields the Answer F1. The Distractor F1 is computed analogously using Co-F and Pr-F.

The key to interpretation lies in comparing the two F1 values. If both Answer F1 and Distractor F1 are high, it indicates that the model is confident and avoids random guessing. However, to assess whether this confidence is grounded in actual knowledge, one must examine the gap between the two F1 scores. A higher Answer F1 relative to Distractor F1 implies that the model's certainty is predominantly concentrated on correct answers. The ratio of Co-T to Co-F thus reflects the internal distribution between correct concepts and misconceptions within the model.

\begin{table}[ht!]
\centering
\caption{Answer vs. Distractor metric families — Precision, Recall \& F1}
\label{tab:answer_distractor_metrics}
\begin{tabular}{ccl}
\toprule
\multicolumn{3}{c}{\textbf{Answer-metric family}} \\
\midrule
Metric & Formula & Interpretation \\
\midrule
\begin{tabular}[c]{@{}l@{}}Answer Precision (AP)\end{tabular}     
    & Co-T / (Co-T + Co-F)        
    & How often was the model correct when confident? \\
\begin{tabular}[c]{@{}l@{}}Answer Recall (AR)\end{tabular}        
    & Co-T / (Pr-T + Co-T)        
    & Did preference for the correct answer lead to certainty? \\
Answer F1
    & $2 \cdot ( \text{AP} \cdot \text{AR} ) / ( \text{AP} + \text{AR} )$
    & Measures precision and consistency of correct certainty \\
\midrule
\multicolumn{3}{c}{\textbf{Distractor-metric family}} \\
\midrule
Metric & Formula & Interpretation \\
\midrule
\begin{tabular}[c]{@{}l@{}}Distractor Precision (DP)\end{tabular} 
    & Co-F / (Co-T + Co-F)        
    & How often was the model wrong when confident? \\
\begin{tabular}[c]{@{}l@{}}Distractor Recall (DR)\end{tabular}    
    & Co-F / (Pr-F + Co-F)        
    & Did preference for a distractor solidify into false certainty? \\
Distractor F1
    & $2 \cdot ( \text{DP} \cdot \text{DR} ) / ( \text{DP} + \text{DR} )$
    & Measures magnitude and consistency of incorrect certainty \\
\bottomrule
\end{tabular}
\end{table}

\subsection{Why this design matters}
The proposed design is structured to eliminate selection bias, block random guessing, and maximize response consistency. By combining these three stages, the framework allows for fine-grained assessment of a model’s true understanding—something that traditional prompt-based evaluations, which often rely heavily on positional bias, may fail to capture. 

In particular, when the values of Co-T and Co-F are sufficiently large, a joint analysis of Answer F1 and Distractor F1 reveals whether the model consistently selects the correct answer for certain topics, or conversely, persistently chooses the same distractor. This enables identification of conceptual gaps and entrenched misconceptions at the level of individual questions. 

The fact that such latent structure can be revealed through the simple procedure of repeated querying makes the proposed method both practical and reproducible as a tool for evaluating LLMs.

\section{Experiments}
\subsection{Datasets \& models}
We conduct evaluation on two benchmarks: Massive Multitask Language Understanding (MMLU, \cite{hendrycks2020measuring}), which tests knowledge-intensive domains, and CommonsenseQA (CSQA, \cite{talmor2019commonsenseqa}), which requires commonsense reasoning. 

The experimental subjects include eight LLMs across four major model families, selected to represent diverse parameter scales and training recipes among both proprietary and open-source systems: ChatGPT (3.5-turbo, 4o-mini) \cite{achiam2023gpt}, Claude (3-haiku \cite{AnthropicClaude32024}, 3.5-sonnet \cite{AnthropicClaude35Add2025}), Gemini (1.5-flash, 1.5-pro) \cite{team2024gemini}, and LLaMA (3-8b, 3-70b) \cite{MetaLlama32024}.

Across all combinations of models and datasets, we applied a unified repetition protocol and positioning algorithm to collect and analyze approximately 700,000 individual responses. The total number of API calls is reported in Appendix~\ref{app:call_volume}.

\subsection{Main results}
Tables~\ref{tab:claude_response_patterns} and~\ref{tab:answer_distractor_scores} present the evaluation results of Claude 3.5-sonnet on the CSQA benchmark. The top-level metrics, Pr-T and Pr-F, indicate the number of correct and incorrect answers when the model is run only once per question. In contrast, Co-T and Co-F count the number of consistent responses—i.e., identical answers across five repeated trials—that are correct or incorrect, respectively.

Compared to the Baseline, SCOPE maintains the total number of correct answers (Pr-T) while increasing the number of consistent correct answers (Co-T) from 96 to 104, representing an 8\% improvement. Although the number of consistent distractors (Co-F) also increases from 34 to 59, this shift reflects a meaningful behavioral change: once position bias is removed, the model begins to respond more confidently—whether correctly or not—which can be further analyzed through the metric families below.

Looking at the answer-metric and distractor-metric families, we observe that Answer F1 improves from 0.853 to 0.911, a substantial gain of 0.058 points indicating enhanced consistency in correct predictions. While Distractor F1 also increases from 0.147 to 0.217 (a 0.070-point rise), the Answer–Distractor gap remains large, decreasing only slightly from 0.706 to 0.694. This confirms that the model's certainty remains predominantly concentrated on correct answers.

Compared to major competing methods, SCOPE demonstrates superior performance. Majority Voting (MV) achieves the second-highest Answer F1 (0.895) by maximizing Answer Recall, but its Distractor F1 surges to 0.252, reflecting a high rate of consistent incorrect predictions. PriDe and CalibEV show partial improvements but also elevate Distractor F1, compromising precision-recall balance. In contrast, SCOPE achieves the highest Answer F1 while keeping Distractor F1 lower than both MV and CalibEV, striking the most favorable balance: boosting consistent correct answers while suppressing consistent distractors.

In conclusion, SCOPE effectively mitigates selection bias and improves overall performance in Claude 3.5-sonnet. By increasing consistent correct predictions and inducing fewer confident mistakes than competing methods, SCOPE further validates its value as a reliable bias-mitigation framework for real-world settings. Full details are available in Appendix~\ref{app:full_experiments}.

\begin{table}[ht!]
\centering
\caption{Claude 3.5-Sonnet on CSQA: raw counts of Pr-T/F and Co-T/F}
\label{tab:claude_response_patterns}
\begin{tabular}{cc ccc ccc ccc ccc ccc ccc}
\toprule
\multirow{2}{*}{Model} & \multirow{2}{*}{Metric} 
& \multicolumn{2}{c}{Baseline} 
& \multicolumn{2}{c}{CalibEV} 
& \multicolumn{2}{c}{DI} 
& \multicolumn{2}{c}{EC} 
& \multicolumn{2}{c}{MV} 
& \multicolumn{2}{c}{PriDE} 
& \multicolumn{2}{c}{SCOPE} \\
\cmidrule(lr){3-4}
\cmidrule(lr){5-6}
\cmidrule(lr){7-8}
\cmidrule(lr){9-10}
\cmidrule(lr){11-12}
\cmidrule(lr){13-14}
\cmidrule(lr){15-16}
& & T & F & T & F & T & F & T & F & T & F & T & F & T & F \\
\midrule
\multirow{2}{*}{Claude (3.5-sonnet)} 
& Pr & 82  & 47  & 25  & 17  & 153 & 128 & 9   & 5   & 46  & 32  & 67  & 33  & 19  & 24 \\
& Co & 337 & 34  & 401 & 57  & 132 & 87  & 84  & 402 & 389 & 33  & 368 & 32  & 398 & 59 \\
\bottomrule
\end{tabular}
\end{table}

\begin{table}[ht!]
\centering
\caption{Claude 3.5-Sonnet on CSQA: Answer-metric vs. Distractor-metric scores}
\label{tab:answer_distractor_scores}
\begin{tabular}{c ccc ccc}
\toprule
\multirow{2}{*}{Method} 
& \multicolumn{3}{c}{Answer} 
& \multicolumn{3}{c}{Distractor} \\
\cmidrule(lr){2-4} \cmidrule(lr){5-7}
& Precision & Recall & F1 
& Precision & Recall & F1 \\
\midrule
Baseline   & 0.9084 & 0.8043 & 0.8532 & 0.0916 & 0.4198 & 0.1504 \\
CalibEV    & 0.8755 & 0.9413 & 0.9072 & 0.1245 & 0.7703 & 0.2144 \\
DI         & 0.6027 & 0.4632 & 0.5238 & 0.3973 & 0.4047 & 0.4010 \\
EC         & 0.1728 & 0.9032 & 0.2901 & 0.8272 & 0.9877 & 0.9004 \\
MV         & 0.9218 & 0.8943 & 0.9078 & 0.0782 & 0.5077 & 0.1355 \\
PriDE      & 0.9200 & 0.8460 & 0.8814 & 0.0800 & 0.4923 & 0.1376 \\
SCOPE      & 0.8709 & 0.9544 & 0.9107 & 0.1291 & 0.7108 & 0.2185 \\
\bottomrule
\end{tabular}
\end{table}

\subsection{Analysis}
After applying SCOPE, the response pattern of Claude 3.5-sonnet exhibits three notable changes. 

The first is the effect of position bias removal. Even after relocating the answer slot according to the inverse-bias distribution derived from null prompts, the total number of correct responses (Pr-T) remained nearly unchanged. More importantly, the number of consistent correct answers (Co-T)—i.e., instances where the model gave the same correct answer across five repeated trials—increased from 96 to 104, an 8\% improvement. This demonstrates that the model can make consistent decisions based on actual knowledge, even when the correct answer is no longer placed in a preferred position.

The second change concerns the gap between Answer F1 and Distractor F1. Answer F1 rose significantly from 0.853 to 0.911, indicating strengthened certainty in correct answers. Although Distractor F1 also increased from 0.147 to 0.217, the gap between the two metrics remained at approximately 0.69 (narrowing only from 0.706 to 0.694). In other words, while the model occasionally displays confidence in incorrect responses, its certainty remains predominantly directed toward correct answers. This supports the core design of our metric system, wherein the distribution of Co-T and Co-F reflects the model’s internal knowledge-to-misconception ratio—a hypothesis now validated empirically.

Lastly, comparison with competing methods further highlights SCOPE's superiority. Majority Voting (MV) maximized Answer Recall and thus appeared to improve performance, but also caused a surge in Distractor F1 to 0.252, indicating a sharp rise in confident incorrect predictions. PriDe and CalibEV exhibited similar side effects. In contrast, SCOPE achieved a greater increase in Answer F1 than in Distractor F1, resulting in a desirable redistribution of the model's certainty toward correct answers. In short, SCOPE removes positional cues while enhancing model consistency and steering that consistency toward correct answers, thereby empirically demonstrating its effectiveness in mitigating selection bias.

This outcome serves as strong empirical support for the theoretical claims presented earlier—specifically, that the lucky-rate is upper-bounded by $1/n$, and that distance-based distractor dispersion effectively prevents near-miss guesses. These mechanisms prove to be robust even on large-scale commonsense benchmarks.

\subsection{Ablation study}
The ablation study compares three experimental conditions that differ based on whether the IP (Inverse-Positioning) and SS (Semantic-Spread) modules are enabled. The first configuration, \textbf{IP + SS}, represents the full implementation of SCOPE in which answer slots are reallocated according to the inverse-bias distribution and the semantically-similar distractor (SSD) is dispersed using a distance-weighted rule. The second condition, \textbf{$\lnot$IP + SS}, enables only the SS module to assess how well SSD dispersion alone can reduce selection bias without positional reallocation. The third, \textbf{IP + $\lnot$SS}, activates only the IP module to isolate the effect of answer-slot reallocation on performance. This comparison allows us to quantify each module's individual contribution and examine their interaction when applied jointly.

Table~\ref{tab:ablation_metrics} presents the results from evaluating Claude 3.5-Sonnet on the MMLU benchmark, quantifying the impact of the IP and SS modules. In the full \textbf{IP + SS} configuration, the model achieved an Answer F1 of 0.918 and a lucky-rate $\ell$ of 0.004, resulting in a pure skill (F1 − $\ell$) of 0.914. Under the \textbf{IP + $\lnot$SS} setting, the lucky-rate $\ell$ remained unchanged at 0.004, but the Answer F1 dropped sharply to 0.563, reducing the pure skill to 0.559—a 0.355-point decline. This confirms that the SS module significantly improves answer selection through semantic separation, even if it does not further reduce the lucky-rate.

Conversely, in the \textbf{$\lnot$IP + SS} condition, the lucky-rate $\ell$ surged to its theoretical upper bound of 0.25, and the Answer F1 decreased to 0.574, yielding a pure skill score of just 0.324—an additional 0.235-point drop. In summary, approximately 63\% of Claude 3.5-Sonnet’s final performance can be attributed to the IP module and 37\% to the SS module. Only when both modules are jointly activated can the model simultaneously minimize the lucky-rate and achieve peak accuracy. Full ablation results are presented in Appendix~\ref{app:ablation}.

\begin{table}[ht!]
\centering
\caption{Ablation of IP \& SS modules: Answer F1, lucky-hit probability ℓ and pure skill}
\label{tab:ablation_metrics}
\begin{tabular}{c c c c}
\toprule
Condition & Answer F1 & $\ell$ & Pure Skill \\
\midrule
IP + SS            & 0.9182 & 0.0040 & 0.9142 \\
$\lnot$IP + SS     & 0.5701 & 0.2500 & 0.3201 \\
IP + $\lnot$SS     & 0.5633 & 0.0040 & 0.5593 \\
\bottomrule
\end{tabular}
\end{table}

\section{Discussion}
Experimental results show that SCOPE consistently improves Answer F1 across a wide range of model sizes and families—including Claude 3.5-sonnet, ChatGPT 3.5-turbo, and LLaMA 3-70b—while limiting the relative increase in Distractor F1. The most striking case is Claude 3.5-sonnet, where Answer F1 increased from 0.853 to 0.911, while Distractor F1 remained at 0.217, preserving a large gap of 0.694 between the two metrics. This indicates that although some model confidence was redistributed toward distractors, a substantial portion was successfully reallocated to correct answers. The answer-metric and distractor-metric families thus reveal the internal distribution of knowledge and misconceptions within the model, offering interpretable signals—particularly when the gap narrows, which may serve as a warning sign.

The framework’s strengths can be summarized in three points. First, it ensures fairness by fundamentally eliminating label and position cues through label normalization and random shuffling of answer order. Second, by using the inverse of the measured position bias distribution, SCOPE guarantees a theoretical upper bound of $1/n$ on the lucky-rate, thereby enforcing uniform difficulty across all models and datasets. Third, the answer/distractor metric families dissect the model's certainty profile across both axes, exposing risks such as consistent distractor selection (Co-F) that standard accuracy metrics overlook. The framework is also lightweight to implement: the position bias distribution $P$ needs to be estimated only once per model, and the reordering logic is dataset-agnostic and easily transferable across domains.

Nonetheless, several limitations remain. First, repeated calls with null prompts may incur high costs for proprietary API-based models; future work should explore adaptive sampling or low-resolution estimation followed by correction. Second, surface-level biases—such as those related to input length, word frequency, or topic—may persist, suggesting the need for multi-dimensional debiasing techniques. Third, the effectiveness of SSD dispersion depends on embedding quality, which may be weaker in domain-specific contexts. Fourth, on benchmarks such as CSQA with a high proportion of commonsense questions, Distractor F1 occasionally rose by up to 0.07 points. This indicates that models may exhibit high certainty even toward misconceptions, pointing to the importance of incorporating risk-aware calibration techniques in post-processing.

In summary, the proposed framework enhances both fairness and interpretability by enabling fine-grained analysis of consistent certainty in the absence of position and label cues. While the cost of repeated prompting and residual biases remain open challenges, SCOPE is the only method that increases F1 across all tested models while maintaining a lower consistent distractor risk than alternatives like MV and CalibEV. This makes it both academically and practically valuable. Future directions include integrating multi-bias mitigation, domain-adaptive embeddings, and confidence calibration strategies to support a broader range of evaluation scenarios.

\section{Conclusion}
This study proposes SCOPE, a structured evaluation framework designed to suppress selection bias—an often-overlooked factor in multiple-choice assessments of large language models (LLMs). By measuring each model’s inherent position bias and reassigning answer slots according to an inverse-bias distribution, SCOPE prevents models from exploiting positional cues. In parallel, semantically-similar distractors are dispersed probabilistically based on distance, blocking shortcut strategies such as near-miss guessing. Together, these mechanisms promote not just accuracy, but genuine certainty in correct responses. Notably, the introduction of answer/distractor metrics enables fine-grained interpretation of model certainty distributions. Across MMLU and CSQA benchmarks, SCOPE achieved consistent performance improvements and bias mitigation across a wide range of models and settings. Unlike prior calibration methods, it also enhances the interpretability of predictions. However, certain limitations remain. These include the computational cost of repeated null prompts, the persistence of surface-level biases, and the sensitivity of semantic dispersion to embedding quality. Future work may extend SCOPE into a unified evaluation framework by incorporating multi-bias mitigation and domain-specific adaptations to improve overall evaluation reliability. By offering a position- and label-agnostic evaluation structure, SCOPE lays the groundwork for bias-resilient assessment of LLMs—marking a new starting point for fair and trustworthy language model evaluation.

\section*{Reproducibility statement}
We provide all code, data, and configurations necessary to reproduce our experiments. The following information enables readers to replicate the main results presented in this paper.

\begin{itemize}
  \item \textbf{Code Repository}\\
  Complete implementation available at:
  \url{https://github.com/WonjunJeong97/SCOPE}
  
  The repository includes:
  \begin{itemize}
    \item Full pipeline source code (\texttt{src/})
    \item Execution scripts (\texttt{scripts/})
    \item Tutorial notebook (\texttt{notebooks/tutorial.ipynb})
    \item Pre-configured settings (\texttt{configs/default.yaml})
  \end{itemize}
  
  \item \textbf{Quick Reproduction}\\
  Single command to reproduce main results:
  \begin{verbatim}
  python src/main.py --model gpt-3.5-turbo --dataset csqa
  \end{verbatim}
  
  For complete evaluation including ablation study:
  \begin{verbatim}
  python src/main.py --model claude-3-5-sonnet --dataset both --ablation
  \end{verbatim}
  
  \item \textbf{Datasets}\\
  Fixed datasets for reproducibility:
  \begin{itemize}
    \item CSQA: 500 questions from validation set
    \item MMLU: 500 questions from test set
  \end{itemize}
  Pre-processed datasets included in \texttt{data/fixed/} directory.
  
  \item \textbf{Environment}\\
  \begin{itemize}
    \item Python 3.8+ (platform-independent)
    \item Dependencies: \texttt{pip install -r requirements.txt}
    \item All code uses standard Python libraries and APIs
  \end{itemize}
  
  \item \textbf{Key Hyperparameters}\\
  \begin{itemize}
    \item Null prompts: $M = 1{,}000$
    \item Evaluation trials: 5 repetitions per question
    \item Temperature: 1.0
    \item Sentence-BERT: \texttt{all-MiniLM-L6-v2}
    \item Random seed: 42 (fixed throughout)
  \end{itemize}
  
  \item \textbf{API Requirements}\\
  The framework supports multiple LLMs via API:
  \begin{itemize}
    \item OpenAI (GPT-3.5-turbo, GPT-4)
    \item Anthropic (Claude-3-haiku, Claude-3.5-sonnet)
    \item Google (Gemini-1.5-flash/pro)
    \item Groq (LLaMA-3-8b/70b)
  \end{itemize}
  API keys should be set in \texttt{.env} file (see \texttt{.env.example}).
  
  \item \textbf{Expected Results}\\
  With fixed seed, results match paper tables within $\pm$0.001:
  \begin{itemize}
    \item Position bias distributions (Figure 2)
    \item Pr-T/F and Co-T/F metrics (Tables 6-7)
    \item Answer/Distractor F1 scores (Table 7)
    \item Lucky-hit probability $\ell \leq 1/n$
  \end{itemize}
\end{itemize}

In summary, we provide a complete, self-contained implementation with fixed datasets and configurations to ensure exact reproducibility. For questions or issues, please use the GitHub issue tracker.

\bibliographystyle{unsrt}  
\bibliography{references}  

\appendix

\section*{Appendix}
\addcontentsline{toc}{section}{Appendix}

\section{Selection distribution across evaluation conditions}
\label{app:preliminary_condition}

Table~\ref{tab:preliminary_condition} presents a quantitative analysis of how fixed answer labels (L) and fixed option order (F) affect selection bias and model accuracy across four experimental conditions on the MMLU benchmark.

In the Baseline condition, where both label and order are fixed, ChatGPT 3.5-turbo exhibited a KLD of 0.0015 and an accuracy of 0.630, while Gemini 1.5-flash recorded a KLD of 0.0004 and an accuracy of 0.849. In the $\lnot$L + F condition, removing label cues alone led to a lower KLD of 0.0014 and slightly higher accuracy of 0.648 for ChatGPT. Gemini also showed minimal change, with a KLD of 0.0011 and accuracy of 0.846, suggesting that eliminating visual label cues reduces positional skew without harming performance.

In the L + $\lnot$F setting, where option order was randomized while labels remained intact, ChatGPT's KLD rose to 0.0032 with an accuracy of 0.636—nearly identical to its Baseline. Claude 3-haiku showed a similar pattern, with a KLD of 0.0033 and accuracy of 0.646.

Under the fully unbiased condition ($\lnot$L + $\lnot$F), all four models exhibited nearly uniform selection distributions (around 25\% per slot), with KLD values falling below 0.0016. However, accuracies declined: 0.232 for ChatGPT, 0.238 for Claude, 0.255 for Gemini, and 0.264 for LLaMA 3-8B. This drop reflects the removal of previously inflated lucky-rates due to label and position bias, exposing the models' actual knowledge levels. In other words, the observed accuracy decline should not be interpreted as performance degradation, but rather as an accurate measurement of pure skill in the absence of bias.

Overall, Table~\ref{tab:preliminary_condition} demonstrates that a simple preliminary experiment—randomizing labels and option orders—can not only equalize the selection distribution but also reveal the model's unbiased knowledge competence.

\begin{table}[ht!]
\centering
\caption{Overall performance across prompt conditions: selection rate, KLD, and accuracy}
\label{tab:preliminary_condition}
\begin{tabular}{cccccc}
\toprule
Model & Metric & Baseline & $\lnot$L + F & L + $\lnot$F & $\lnot$L + $\lnot$F \\
\midrule
\multirow{3}{*}{\shortstack{ChatGPT\\(3.5-turbo)}}
& Selection Rate & \raisebox{-0.5\height}{\includegraphics[width=0.15\textwidth]{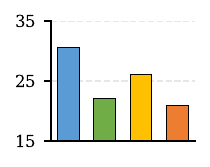}} & \raisebox{-0.5\height}{\includegraphics[width=0.15\textwidth]{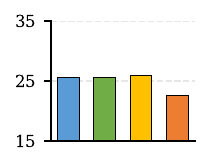}} & \raisebox{-0.5\height}{\includegraphics[width=0.15\textwidth]{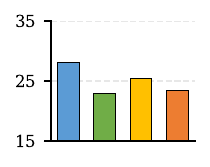}} & \raisebox{-0.5\height}{\includegraphics[width=0.15\textwidth]{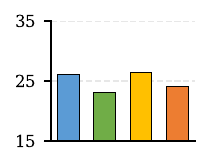}} \\
& KLD   & 0.0015 & 0.0014 & 0.0032 & 0.0016 \\
& Accuracy & 0.6300 & 0.6480 & 0.6360 & 0.2320 \\
\midrule
\multirow{3}{*}{\shortstack{Claude\\(3-haiku)}} 
& Selection Rate & \raisebox{-0.5\height}{\includegraphics[width=0.15\textwidth]{figure/app.A/claude_baseline.pdf}} & \raisebox{-0.5\height}{\includegraphics[width=0.15\textwidth]{figure/app.A/claude_not_L+F.pdf}} & \raisebox{-0.5\height}{\includegraphics[width=0.15\textwidth]{figure/app.A/claude_L+not_F.pdf}} & \raisebox{-0.5\height}{\includegraphics[width=0.15\textwidth]{figure/app.A/claude_not_L+not_F.pdf}} \\
& KLD   & 0.0191 & 0.0101 & 0.0033 & 0.0018 \\
& Accuracy & 0.6760 & 0.6710 & 0.6460 & 0.2380 \\
\midrule
\multirow{3}{*}{\shortstack{Gemini\\(1.5-flash)}} 
& Selection Rate & \raisebox{-0.5\height}{\includegraphics[width=0.15\textwidth]{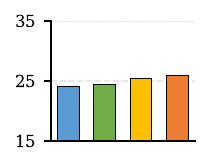}} & \raisebox{-0.5\height}{\includegraphics[width=0.15\textwidth]{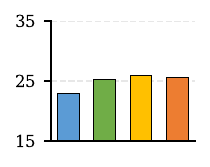}} & \raisebox{-0.5\height}{\includegraphics[width=0.15\textwidth]{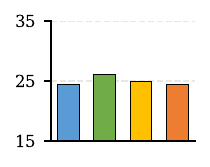}} & \raisebox{-0.5\height}{\includegraphics[width=0.15\textwidth]{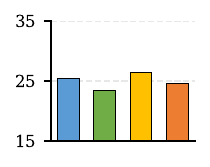}} \\
& KLD   & 0.0004 & 0.0011 & 0.0004 & 0.0010 \\
& Accuracy & 0.8490 & 0.8460 & 0.8380 & 0.2550 \\
\midrule
\multirow{3}{*}{\shortstack{LLaMA\\(3-8b)}} 
& Selection Rate & \raisebox{-0.5\height}{\includegraphics[width=0.15\textwidth]{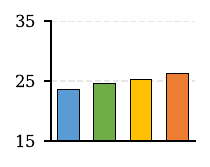}} & \raisebox{-0.5\height}{\includegraphics[width=0.15\textwidth]{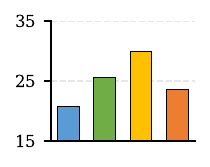}} & \raisebox{-0.5\height}{\includegraphics[width=0.15\textwidth]{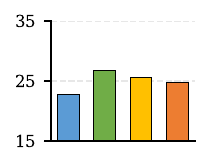}} & \raisebox{-0.5\height}{\includegraphics[width=0.15\textwidth]{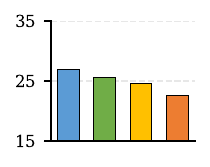}} \\
& KLD   & 0.0007 & 0.0087 & 0.0017 & 0.0019 \\
& Accuracy & 0.6700 & 0.6180 & 0.6230 & 0.2640 \\
\bottomrule
\end{tabular}
\end{table}

\section{Accuracy impact of low-bias re-positioning}
\label{app:preliminary_lbp}

Table~\ref{tab:preliminary_lbp} reports the impact of the Low-Bias Positioning (LBP) condition on model performance using the MMLU benchmark. Under this setting, the correct answer is placed in the position least preferred by the model under Baseline measurements.

Compared to the Baseline, accuracy dropped by 0.020 points for ChatGPT 3.5-turbo (from 0.6330 to 0.6130) and by a substantial 0.137 points for Claude 3-haiku (from 0.6700 to 0.5330). Gemini 1.5-flash and LLaMA 3-8B also showed decreases of 0.033 and 0.061 points, respectively.

The selection rate distribution between Baseline and LBP conditions is completely reversed, as the correct answer is deliberately placed in the model's least-preferred position. Despite using the same questions, the consistent drop in accuracy indicates that model performance is strongly influenced by the location of the correct answer. This also highlights a key limitation of simple randomization: randomly shuffling the answer order does not fully eliminate selection bias, since random placements may still coincide with positions favored by the model.

This preliminary LBP experiment underscores that, without proper control of position bias, model performance can be easily inflated. It further emphasizes the need for additional correction mechanisms to ensure fair evaluation.

\setlength{\tabcolsep}{3pt}
\begin{table}[ht!]
  \centering
  \caption{Selection rate and accuracy under baseline vs. low-bias placement (LBP)}
  \label{tab:preliminary_lbp}

  \begin{subtable}[t]{0.48\linewidth}
    \centering
    \caption{ChatGPT (3.5-turbo)}
    \label{tab:lbp_chatgpt}
    \vspace{0.4em}
    \begin{tabular}{ccc}
      \toprule
      Metric & Baseline & LBP \\
      \midrule
      Selection Rate &
        \raisebox{-0.5\height}{\includegraphics[width=.23\linewidth]{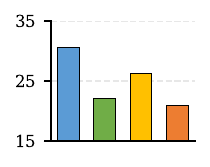}} &
        \raisebox{-0.5\height}{\includegraphics[width=.23\linewidth]{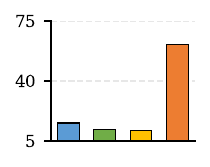}} \\
      Accuracy & 0.6330 & 0.6130 \\
      \bottomrule
    \end{tabular}
  \end{subtable}
  \hfill
  \begin{subtable}[t]{0.48\linewidth}
    \centering
    \caption{Claude (3-haiku)}
    \label{tab:lbp_claude}
    \vspace{0.4em}
    \begin{tabular}{ccc}
      \toprule
      Metric & Baseline & LBP \\
      \midrule
      Selection Rate &
        \raisebox{-0.5\height}{\includegraphics[width=.23\linewidth]{figure/app.B/claude_baseline.pdf}} &
        \raisebox{-0.5\height}{\includegraphics[width=.23\linewidth]{figure/app.B/claude_lbp.pdf}} \\
      Accuracy & 0.6700 & 0.5330 \\
      \bottomrule
    \end{tabular}
  \end{subtable}
  \vspace{0.4em}

  \begin{subtable}[t]{0.48\linewidth}
    \centering
    \caption{Gemini (1.5-flash)}
    \label{tab:lbp_gemini}
    \vspace{0.4em}
    \begin{tabular}{ccc}
      \toprule
      Metric & Baseline & LBP \\
      \midrule
      Selection Rate &
        \raisebox{-0.5\height}{\includegraphics[width=.23\linewidth]{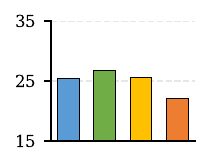}} &
        \raisebox{-0.5\height}{\includegraphics[width=.23\linewidth]{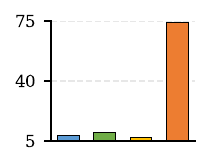}} \\
      Accuracy & 0.7770 & 0.7440 \\
      \bottomrule
    \end{tabular}
  \end{subtable}
  \hfill
  \begin{subtable}[t]{0.48\linewidth}
    \centering
    \caption{LLaMA (3-8B)}
    \label{tab:lbp_llama}
    \vspace{0.4em}
    \begin{tabular}{ccc}
      \toprule
      Metric & Baseline & LBP \\
      \midrule
      Selection Rate &
        \raisebox{-0.5\height}{\includegraphics[width=.23\linewidth]{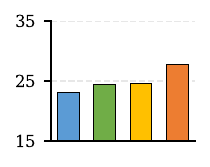}} &
        \raisebox{-0.5\height}{\includegraphics[width=.23\linewidth]{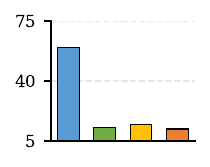}} \\
      Accuracy & 0.6600 & 0.5990 \\
      \bottomrule
    \end{tabular}
  \end{subtable}
\end{table}

\section{Adjacent-placement condition}
\label{app:preliminary_ssd}

Table~\ref{tab:preliminary_ssd} illustrates how the placement of a semantically-similar distractor (SSD)—whether adjacent to or distant from the correct answer—affects LLM performance.

For Claude 3-haiku, when SSD adjacency was prohibited ($\lnot$A), the SSD selection rate dropped sharply by 8.8 percentage points, while accuracy decreased by only 0.7 points. This suggests that Claude makes decisions primarily based on semantic content, and that once the SSD is no longer in close proximity, near-miss guessing becomes rare. In other words, even when the SSD is placed far away, Claude still identifies the correct answer reliably, free from the misleading influence of a tempting distractor.

In contrast, LLaMA 3-8B exhibited the opposite trend. When SSDs were moved away, the SSD selection rate increased by 5.8 points and accuracy fell by 3.7 points. This implies that LLaMA relies heavily on superficial cues such as position or label, rather than semantic understanding. When the SSD is adjacent to the correct answer, LLaMA may exploit positional hints to maximize accuracy, potentially compromising the fairness of evaluation.

ChatGPT 3.5-turbo and Gemini 1.5-flash fall somewhere in between. For both models, changes in SSD selection rate and accuracy were within 2 percentage points under the $\lnot$A condition. This intermediate behavior suggests that these models are neither as semantically grounded as Claude nor as position-biased as LLaMA. In such cases, SSD separation serves more as a tool for modulating evaluation difficulty or as a diagnostic indicator for positional sensitivity.

In summary, distancing the SSD from the correct answer sharply reveals each model's attentional bias. For semantically driven models like Claude, it acts as a clear mechanism to suppress near-miss guessing. For position-sensitive models like LLaMA, it serves as a safeguard against unfair performance inflation. This study provides the first quantitative account of such behavior, and future work should explore more granular SSD placement strategies tailored to specific model architectures.

\begin{table}[ht!]
\centering
\caption{SSD selection rate and accuracy for adjacent versus non-adjacent placement}
\label{tab:preliminary_ssd}
\begin{tabular}{ccccc}
\toprule
Model & Metric & Baseline & A & $\lnot$A \\
\midrule
\multirow{2}{*}{\shortstack{ChatGPT (3.5-turbo)}} 
  & SSD selection rate & 0.3770 & 0.3550 & 0.3830 \\
  & Accuracy  & 0.6390 & 0.6310 & 0.6400 \\
\midrule
\multirow{2}{*}{\shortstack{Claude (3-haiku)}} 
  & SSD selection rate & 0.4220 & 0.4360 & 0.3340 \\
  & Accuracy  & 0.6780 & 0.6700 & 0.6710 \\
\midrule
\multirow{2}{*}{\shortstack{Gemini (1.5-flash)}} 
  & SSD selection rate & 0.3820 & 0.3660 & 0.3770 \\
  & Accuracy  & 0.8480 & 0.8470 & 0.8250 \\
\midrule
\multirow{2}{*}{\shortstack{LLaMA (3-8b)}} 
  & SSD selection rate & 0.3860 & 0.3910 & 0.4440 \\
  & Accuracy  & 0.6790 & 0.6830 & 0.6420 \\
\bottomrule
\end{tabular}
\end{table}

\section{Reference algorithms (IP \& SS)}

\subsection{Inverse-positioning (IP) module}
Algorithm~\ref{alg:ip} estimates the model's position bias distribution $\mathbf{P} = (p_1, \dots, p_n)$ by invoking the null prompt $u$ a total of $M$ times and accumulating the selection frequency $c_i$ for each answer slot. Laplace smoothing with parameter $\varepsilon$ is then applied to stabilize the estimate. Next, the inverse-bias distribution $\mathbf{Q}$ is computed by setting $q_i \propto 1/p_i$ and normalizing. This results in a reweighting scheme where positions with higher original bias are assigned lower selection probability. Through this mechanism, SCOPE deliberately allocates correct answers to less preferred positions.

The computational complexity of IP is $O(M + n)$, and the module requires only two hyperparameters: the number of queries $M$ and the smoothing constant $\varepsilon$. For $M \ge 100$, a small value such as $\varepsilon \approx 10^{-3}$ is generally sufficient; for very small $M$, it is safe to increase $\varepsilon$ to a range between 0.5 and 1.

\begin{algorithm}[ht!]
\DontPrintSemicolon
\caption{IP(Inverse-Positioning) module}\label{alg:ip}

\SetKwInput{KwIn}{Input}
\SetKwInput{KwOut}{Output}

\KwIn{LLM $f$; number of options $n$;
      Null-prompt budget $M$;
      Neutral-prompt template $u$;
      $\varepsilon$ $(\!\ll 1)$}
\KwOut{Inverse-bias distribution $\mathbf{Q} = (q_1,\ldots,q_n)$}

\BlankLine
\For{$i \leftarrow 1$ \KwTo $n$}{ $c_i \leftarrow 0$ \tcp*[r]{init counts}
}

\For{$t \leftarrow 1$ \KwTo $M$}{
  $k \leftarrow f(u)$\tcp*[r]{LLM returns index $k \in \{1,\dots,n\}$}
  $c_k \leftarrow c_k + 1$
}

\For{$i \leftarrow 1$ \KwTo $n$}{
  $p_i \leftarrow \dfrac{c_i + \varepsilon}{\,M + n\varepsilon\,}$
}

$\displaystyle q_i \leftarrow
 \dfrac{1 / p_i}{\sum_{j=1}^{n} 1 / p_j}\quad
 (i = 1,\dots,n)$\tcp*[r]{normalize inverse bias}

\KwRet{$\mathbf{Q}$}
\end{algorithm}

\subsection{Semantic-spread (SS) module}
Algorithm~\ref{alg:ss} first samples a new answer slot $i^\star$ from the inverse-bias distribution $\mathbf{Q}$ generated by the IP module. Using Sentence-BERT embeddings $\phi$, it identifies the distractor $d$ with the highest cosine similarity to the correct option $o_a$, designating it as the semantically-similar distractor (SSD). Then, a probability vector $\mathbf{R}$ is constructed using distance-based weights $w_s = |s - i^\star|^{\tau}$, and the SSD's final position $j_{\text{SSD}}$ is resampled from $\mathbf{R}$ to maximize its distance from the correct answer.

A higher $\tau$ enforces stronger separation between the answer and SSD; when $\tau = 0$, the distribution converges to uniform. The SS module operates with computational complexity $O(n)$ and requires no model queries, making it efficient and reproducible once the hyperparameters $\tau$ and embedding model $\phi$ are fixed.

These two modules are invoked as abstract procedures in Algorithm~\ref{alg:scope} of the main text. The IP module provides $\mathbf{Q}$, and the SS module returns the pair $\langle i^\star, j_{\text{SSD}} \rangle$, enabling SCOPE to jointly suppress position bias and near-miss guessing caused by semantic proximity.

\begin{algorithm}[ht!]
\DontPrintSemicolon
\caption{SS(Semantic-Spread) Module}\label{alg:ss}

\SetKwInput{KwIn}{Input}
\SetKwInput{KwOut}{Output}

\KwIn{Option set $\mathcal{O}=\{o_1,\dots,o_n\}$;
      Ground truth index $a$;
      Inverse bias distribution $\mathbf{Q}$;
      Sentence-BERT encoder $\phi$;
      Temperature $\tau>0$}
\KwOut{$\langle i^\star,\;j_{\text{SSD}}\rangle$}

\BlankLine
$i^\star \sim \mathbf{Q}$\;

\For{$k \leftarrow 1$ \KwTo $n$}{
$\mathbf e_k \gets \phi(o_k)$ \tcp*{embed all options}
}
$\displaystyle d \leftarrow
  \arg\max_{k\neq a}
  \frac{\mathbf e_a^{\!\top}\mathbf e_k}
       {\lVert\mathbf e_a\rVert\;\lVert\mathbf e_k\rVert}$
\tcp*[r]{SSD index}

\For{$s\leftarrow 1$ \KwTo $n$}{
  \eIf{$s = i^\star$}{$w_s \gets 0$}
       {$w_s \gets |s-i^\star|^{\tau}$}
}
$\displaystyle
  r_s \;\gets\; \frac{w_s}{\sum_{t=1}^{n} w_t}
  \quad (s=1,\dots,n)$
\tcp*[r]{normalize $\mathbf w \to \mathbf R$}

$j_{\text{SSD}} \sim \mathbf{R}$

\KwRet{$\langle i^\star,\;j_{\text{SSD}}\rangle$}
\end{algorithm}

\section{Proof details: Position-bias cancellation}
\label{app:proof_bias_cancel}
This section provides full derivations of the theorems presented in the main text, with all intermediate steps shown in detail.

\subsection*{Harmonic–arithmetic mean absolute inequality}
For any set of positive real numbers $x_1, \dots, x_n$, the following inequality holds:
\[
 \frac{n}{\sum_{j=1}^{n} 1/x_j} \leq \frac{\sum_{j=1}^{n} x_j}{n}, \qquad (\textbf{HM} \leq \textbf{AM})
\]
Equality holds if and only if all $x_j$ are equal.

\subsection*{Variable definitions}
\begin{itemize}
    \item Position bias distribution: $P = (p_1, \dots, p_n), \quad p_i > 0, \quad \sum_{i} p_i = 1$
    \item Inverse-bias distribution: $q_i = \frac{1/p_i}{\sum_{k=1}^{n}1/p_k}, \qquad Q = (q_1, \dots, q_n)$
    \item Answer slot: $i^* \sim Q$
    \item Lucky-rate: $\ell = \textbf{Prob(model selects correct answer based on position only)} = \sum_{i=1}^{n} p_i q_i$
\end{itemize}

\subsection*{Closed-form luck probability}
\begin{equation}
\label{eq:num_1}
\ell = \frac{\sum_{i=1}^{n} p_i \cdot \frac{1}{p_i}}{\sum_{k=1}^{n}\frac{1}{p_k}} = \frac{n}{\sum_{k=1}^{n} \frac{1}{p_k}}
\end{equation}

\textit{Explanation:} In the numerator, $p_i$ cancels out with $1/p_i$ for each term, resulting in a constant sum of $n$. Therefore, the value of $\ell$ is entirely determined by the denominator $\sum 1/p_k$.

\subsection*{Upper-bound derivation}
To apply the harmonic–arithmetic mean inequality (HM $\leq$ AM) to Equation~\ref{eq:num_1}, let $x_k = 1/p_k$. Then the inequality becomes:

\begin{equation}
\label{eq:num_2}
\sum_{k=1}^{n} \frac{1}{p_k} \geq n^2
\end{equation}

Equality holds if and only if $p_1 = \dots = p_n = 1/n$. Substituting Equation~\ref{eq:num_2} into Equation~\ref{eq:num_1} yields:

\begin{equation}
\label{eq:num_3}
\ell \leq \frac{n}{n^2} = \frac{1}{n}
\end{equation}

\subsection*{Lower-bound derivation}
Let $p_{\min} = \min_i p_i$ denote the smallest position probability.

Since all $p_k \geq p_{\min}$, it follows that:

\begin{equation}
\label{eq:num_4}
\sum_{k=1}^{n} \frac{1}{p_k} \leq \frac{n}{p_{\min}}
\end{equation}

Substituting Equation~\ref{eq:num_4} into Equation~\ref{eq:num_1} gives:

\begin{equation}
\label{eq:num_5}
\ell \geq p_{\min}
\end{equation}

\subsection*{Conclusion}
In summary,
\[
p_{\min} \leq \ell \leq \frac{1}{n}
\]

\begin{itemize}
    \item Upper bound equality: When $P$ is uniform $(p_i = 1/n)$, we have $\ell = 1/n$.
    \item Lower bound equality: When one position has probability 1 and all others 0, then $\ell = p_{\min} = 0$. (However, since $p_i > 0$ by definition, this bound is only approached in the limit $\displaystyle \lim_{p_{\min} \to 0}$.)
\end{itemize}

Thus, inverse-bias positioning strictly bounds the model’s expected advantage from position bias below that of random guessing.

\section{Proof details: Distance-weighted distractor dispersion}
\label{app:proof_SSD}

\subsection*{Problem setting}
Let $i^*$ denote the position of the correct answer. Define the set of distractor positions as $S = \left\{1, \dots, n \right\} \setminus \left\{i^*\right\}$, and let the distance to the correct answer for each $j \in S$ be given by $d_j = \left| i^* - j \right|$.

\textbf{Expected distance under uniform distribution:}
\[
\mu_\textbf{unif} = \frac{1}{n-1} \sum_{j \in S} d_j
\]

\textbf{Expected distance under exponential-weighted distribution $R$:} Let $w_j = e^{d_j}$ and $r_j = \frac{w_j}{\sum_{k \in S} w_k}$. Then the expected distance becomes:
\[
\mu_R = \sum_{j \in S} r_j d_j
\]

\subsection*{Reformulation for Jensen’s inequality}
Let $\varphi(x) = e^x$ be a convex function and define uniform probability mass $\widetilde{r}_j = \frac{1}{n-1}$. Then:

\[
\mu_\textbf{unif} = \sum_{j} \widetilde{r}_j d_j, \qquad \sum_{j} \widetilde{r}_j \varphi(d_j) = \frac{1}{n-1} \sum_{j} e^{d_j}
\]

Applying Jensen’s inequality:

\[
\varphi \left( \sum_{j} \widetilde{r}_j d_j \right) \leq \sum_{j} \widetilde{r}_j \varphi(d_j)
\]

yields:

\begin{equation}
\label{eq:num_6}
e^{\mu_\textbf{unif}} \leq \frac{1}{n-1} \sum_{j} e^{d_j}
\end{equation}

\subsection{Exponential-weighted expected distance}
Under the exponential-weighted distribution, we have:

\begin{equation}
\label{eq:num_7}
\mu_R = \frac{\sum_{j} e^{d_j} d_j}{\sum_{j} e^{d_j}} 
= \frac{d_{\max} \sum_{j} e^{d_j} - \sum_{j} e^{d_j} (d_{\max} - d_j)}{\sum_{j} e^{d_j}} 
\geq d_{\max} - \frac{\sum_{j} e^{d_j} (d_{\max} - d_j)}{\sum_{j} e^{d_j}}
\end{equation}

Here, $d_{\max} = \max_{j \in S} d_j \geq 1$. The second term is a weighted average, so it satisfies $0 < \cdot < d_{\max} - 1$.

Therefore,

\begin{equation}
\label{eq:num_8}
\mu_R > d_{\max} - (d_{\max} - 1) = 1
\end{equation}

\subsection{Uniform expected-distance upper bound}
Since the set of distances includes at least 0, 1, or larger values, we have:

\begin{equation}
\label{eq:num_9}
\mu_\textbf{unif} \leq \frac{(n-2) d_{\max}}{n-1}
\end{equation}

When $n \geq 3$, the right-hand side is strictly less than $d_{\max}$. In particular, since $d_{\max} \geq 1$, it follows that:

\begin{equation}
\label{eq:num_10}
\mu_\textbf{unif} \leq d_{\max} - \frac{d_{\max} - 1}{n-1} < d_{\max} \leq \mu_R
\end{equation}

\subsection{Conclusion}
Combining Equations~\ref{eq:num_8}, \ref{eq:num_9}, and \ref{eq:num_10}, we conclude:

\[
\mu_R > \mu_\textbf{unif}, \qquad \text{for } n \geq 3
\]

The exponential-weighted distribution systematically spreads distractors farther from the correct answer, thereby always increasing the expected distance. When $n = 2$, the set $S$ contains only one position, so both expectations equal 1: $\mu_R = \mu_\textbf{unif} = 1$. In this case, the inequality does not hold.

\section{Experimental call-volume budget}
\label{app:call_volume}
The call volume used in our experiments is calculated based on a core unit in which each LLM is evaluated under identical settings across 500 randomly sampled items from each dataset (MMLU and CSQA), with five repetitions per item. For a single method, this amounts to $2 \times 500 \times 5 \times 8 = 40{,}000$ calls across two datasets, five repetitions, and eight models. Depending on the evaluation method, additional operations may increase the total call count. The table below summarizes the multiplier and estimated total call volume required by each method.

\begin{table}[ht!]
\centering
\caption{Additional call overhead and total call volume by evaluation method}
\label{tab:total_calls}
\begin{tabular}{c c c c}
\toprule
Method & Additional Operation & Multiplier & Total Calls \\
\midrule
Baseline & None                     & $\times$1     & 40k \\
CalibEV  & Post-processing only     & $\times$1     & 40k \\
DI       & Prompt injection         & $\times$1     & 40k \\
EC       & Generate reasoning step & $\times$2     & 80k \\
MV       & 10 randomized permutations & $\times$10    & 400k \\
PriDe    & Prior estimation          & $\times$1.05  & 42k \\
SCOPE    & Null prompt-based adjustment & +8k      & 48k \\
\midrule
Total    &                          &               & 690k \\
\bottomrule
\end{tabular}
\end{table}

\section{Empirical position bias \& counter-bias distributions}
\label{app:bias_dists}

\subsection{4-option multiple-choice}
Table~\ref{tab:position_bias_4choices} displays the observed position bias and inverse-bias distributions for eight model versions on 4-option multiple-choice items. 

In terms of position bias, the ChatGPT series, LLaMA series, and Gemini 1.5-flash tended to select the first option most frequently. In contrast, Claude 3-haiku and Gemini 1.5-pro showed a preference for the last option. Notably, both Claude 3-haiku and Gemini 1.5-pro also exhibited a marked bias toward the third option, which is adjacent to the final slot.

\setlength{\tabcolsep}{3pt}
\begin{table}[ht!]
  \centering
  \caption{Position and inverse-bias distributions in 4-choice questions}
  \label{tab:position_bias_4choices}

  \begin{subtable}[t]{0.48\linewidth}
    \centering
    \caption{ChatGPT}
    \label{tab:posbias_chatgpt_mmlu}
    \vspace{0.4em}
    \begin{tabular}{ccc}
      \toprule
      Metric & 3.5-turbo & 4o-mini \\
      \midrule
      Position Bias &
        \raisebox{-0.5\height}{\includegraphics[width=.26\linewidth]{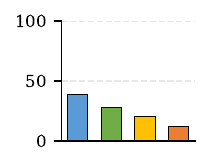}} &
        \raisebox{-0.5\height}{\includegraphics[width=.26\linewidth]{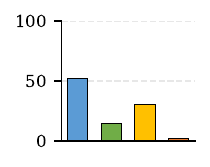}} \\
      Inverse Bias  &
        \raisebox{-0.5\height}{\includegraphics[width=.26\linewidth]{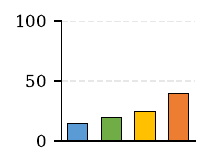}} &
        \raisebox{-0.5\height}{\includegraphics[width=.26\linewidth]{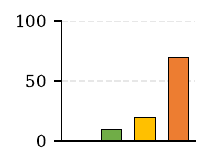}} \\
      \bottomrule
    \end{tabular}
  \end{subtable}
  \hfill
  \begin{subtable}[t]{0.48\linewidth}
    \centering
    \caption{Claude}
    \label{tab:posbias_claude_mmlu}
    \vspace{0.4em}
    \begin{tabular}{ccc}
      \toprule
      Metric & 3-haiku & 3.5-sonnet \\
      \midrule
      Position Bias &
        \raisebox{-0.5\height}{\includegraphics[width=.26\linewidth]{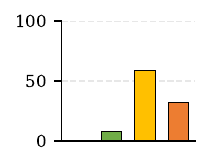}} &
        \raisebox{-0.5\height}{\includegraphics[width=.26\linewidth]{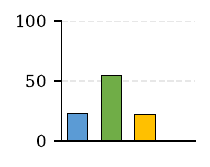}} \\
      Inverse Bias  &
        \raisebox{-0.5\height}{\includegraphics[width=.26\linewidth]{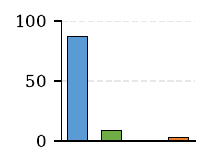}} &
        \raisebox{-0.5\height}{\includegraphics[width=.26\linewidth]{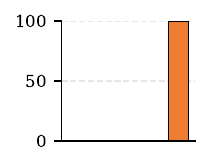}} \\
      \bottomrule
    \end{tabular}
  \end{subtable}
  \vspace{0.4em}

  \begin{subtable}[t]{0.48\linewidth}
    \centering
    \caption{Gemini}
    \label{tab:posbias_gemini_mmlu}
    \vspace{0.4em}
    \begin{tabular}{ccc}
      \toprule
      Metric & 1.5-flash & 1.5-pro \\
      \midrule
      Position Bias &
        \raisebox{-0.5\height}{\includegraphics[width=.26\linewidth]{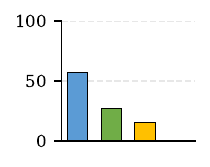}} &
        \raisebox{-0.5\height}{\includegraphics[width=.26\linewidth]{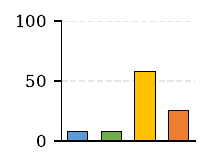}} \\
      Inverse Bias  &
        \raisebox{-0.5\height}{\includegraphics[width=.26\linewidth]{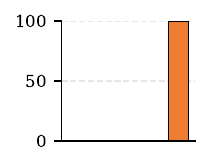}} &
        \raisebox{-0.5\height}{\includegraphics[width=.26\linewidth]{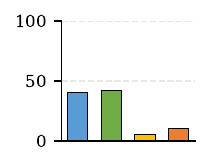}} \\
      \bottomrule
    \end{tabular}
  \end{subtable}
  \hfill
  \begin{subtable}[t]{0.48\linewidth}
    \centering
    \caption{LLaMA}
    \label{tab:posbias_llama_mmlu}
    \vspace{0.4em}
    \begin{tabular}{ccc}
      \toprule
      Metric & 3-8B & 3-70B \\
      \midrule
      Position Bias &
        \raisebox{-0.5\height}{\includegraphics[width=.26\linewidth]{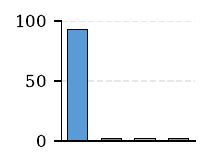}} &
        \raisebox{-0.5\height}{\includegraphics[width=.26\linewidth]{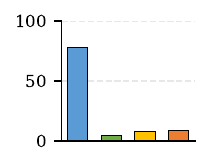}} \\
      Inverse Bias  &
        \raisebox{-0.5\height}{\includegraphics[width=.26\linewidth]{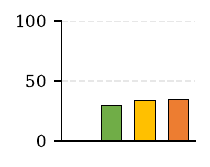}} &
        \raisebox{-0.5\height}{\includegraphics[width=.26\linewidth]{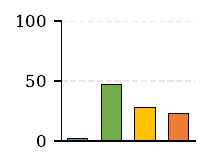}} \\
      \bottomrule
    \end{tabular}
  \end{subtable}
\end{table}

\subsection{5-option multiple-choice}
Table~\ref{tab:position_bias_5choices} summarizes the observed position bias and inverse-bias distributions for each model under the 5-option multiple-choice setting. ChatGPT 4o-mini, Gemini 1.5-flash, and the LLaMA models exhibited a strong primacy effect, showing a tendency to favor the first option. In contrast, ChatGPT 3.5-turbo and Claude 3-haiku revealed a recency effect, unconsciously prioritizing the fourth or final option. Meanwhile, Gemini 1.5-pro and Claude 3.5-sonnet showed relatively higher selection rates for middle positions, suggesting that—similar to the 4-option results—they are comparatively less affected by serial position effects.

These observations across Table~\ref{tab:position_bias_4choices} and Table~\ref{tab:position_bias_5choices} are strikingly similar to the well-documented Serial Position Effect in cognitive psychology \cite{murdock1962serial, glanzer1966two}, which states that items at the beginning (primacy) and end (recency) of a list are cognitively prioritized. Large language models, like humans, appear to unconsciously favor the first or last options in a list, which may allow them to achieve artificially high scores irrespective of true comprehension. This not only undermines fairness in evaluation metrics and reduces the reliability of model comparisons but also suggests that internal mechanisms such as tokenizers and decoders may overencode positional cues.

Recent studies (see Section~\ref{subsec:cognitive_bias}) have shown that human cognitive biases can manifest in LLMs, and psychological techniques have been partially adopted to mitigate such effects. Incorporating these approaches into LLM evaluation is highly desirable. Future research should therefore explore a broader set of cognitive debiasing strategies to counteract the Serial Position Effect in model assessments.

\setlength{\tabcolsep}{3pt}
\begin{table}[ht!]
  \centering
  \caption{Position and inverse-bias distributions in 5-choice questions}
  \label{tab:position_bias_5choices}

  \begin{subtable}[t]{0.48\linewidth}
    \centering
    \caption{ChatGPT}
    \label{tab:posbias_chatgpt}
    \vspace{0.4em}
    \begin{tabular}{ccc}
      \toprule
      Metric & 3.5-turbo & 4o-mini \\
      \midrule
      Position Bias &
        \raisebox{-0.5\height}{\includegraphics[width=.26\linewidth]{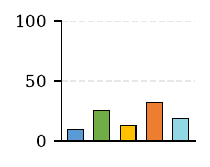}} &
        \raisebox{-0.5\height}{\includegraphics[width=.26\linewidth]{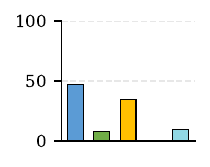}} \\
      Inverse Bias  &
        \raisebox{-0.5\height}{\includegraphics[width=.26\linewidth]{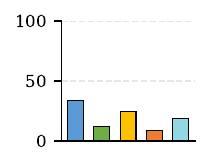}} &
        \raisebox{-0.5\height}{\includegraphics[width=.26\linewidth]{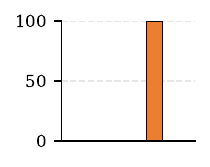}} \\
      \bottomrule
    \end{tabular}
  \end{subtable}
  \hfill
  \begin{subtable}[t]{0.48\linewidth}
    \centering
    \caption{Claude}
    \label{tab:posbias_claude}
    \vspace{0.4em}
    \begin{tabular}{ccc}
      \toprule
      Metric & 3-haiku & 3.5-sonnet \\
      \midrule
      Position Bias &
        \raisebox{-0.5\height}{\includegraphics[width=.26\linewidth]{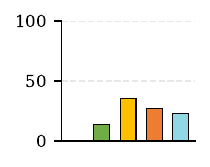}} &
        \raisebox{-0.5\height}{\includegraphics[width=.26\linewidth]{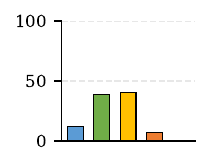}} \\
      Inverse Bias  &
        \raisebox{-0.5\height}{\includegraphics[width=.26\linewidth]{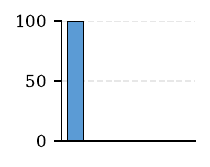}} &
        \raisebox{-0.5\height}{\includegraphics[width=.26\linewidth]{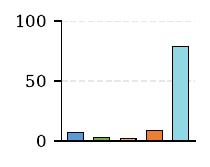}} \\
      \bottomrule
    \end{tabular}
  \end{subtable}
  \vspace{0.4em}
  
  \begin{subtable}[t]{0.48\linewidth}
    \centering
    \caption{Gemini}
    \label{tab:posbias_gemini}
    \vspace{0.4em}
    \begin{tabular}{ccc}
      \toprule
      Metric & 1.5-flash & 1.5-pro \\
      \midrule
      Position Bias &
        \raisebox{-0.5\height}{\includegraphics[width=.26\linewidth]{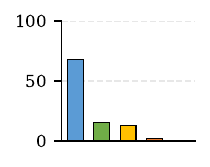}} &
        \raisebox{-0.5\height}{\includegraphics[width=.26\linewidth]{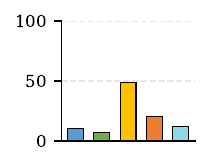}} \\
      Inverse Bias  &
        \raisebox{-0.5\height}{\includegraphics[width=.26\linewidth]{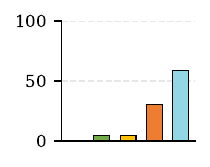}} &
        \raisebox{-0.5\height}{\includegraphics[width=.26\linewidth]{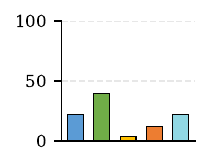}} \\
      \bottomrule
    \end{tabular}
  \end{subtable}
  \hfill
  \begin{subtable}[t]{0.48\linewidth}
    \centering
    \caption{LLaMA}
    \label{tab:posbias_llama}
    \vspace{0.4em}
    \begin{tabular}{ccc}
      \toprule
      Metric & 3-8B & 3-70B \\
      \midrule
      Position Bias &
        \raisebox{-0.5\height}{\includegraphics[width=.26\linewidth]{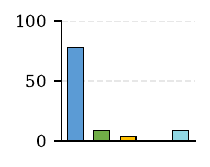}} &
        \raisebox{-0.5\height}{\includegraphics[width=.26\linewidth]{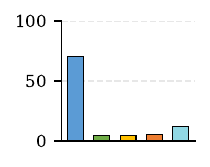}} \\
      Inverse Bias  &
        \raisebox{-0.5\height}{\includegraphics[width=.26\linewidth]{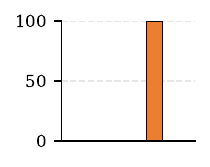}} &
        \raisebox{-0.5\height}{\includegraphics[width=.26\linewidth]{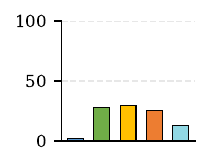}} \\
      \bottomrule
    \end{tabular}
  \end{subtable}
\end{table}

\section{Complete benchmark results}
\label{app:full_experiments}

\subsection{MMLU benchmark results}
Table~\ref{tab:Pr_Co_MMLU} reports the raw counts of Pr-T/F and Co-T/F for each model-method combination, while Tables~\ref{tab:ChatGPT_eval_MMLU}, \ref{tab:Claude_eval_MMLU}, \ref{tab:Gemini_eval_MMLU}, and \ref{tab:LLaMA_eval_MMLU} summarize the Answer and Distractor metric families per model. Taken together, several clear trends emerge.

Across the MMLU benchmark, SCOPE consistently achieved performance improvements while mitigating selection bias and strengthening overall confidence in correct answers. Specifically, Answer F1 increased for all eight LLMs by at least 0.01 points and up to 0.12 points. Notably, SCOPE achieved the highest F1 among all methods for ChatGPT 4o-mini and Gemini 1.5-flash, and remained within 0.002 points of the top-performing method for the remaining six models. Unlike other techniques that exhibit superiority only for specific model types, SCOPE demonstrated robust gains regardless of model size or architecture.

Moreover, SCOPE simultaneously mitigates two distinct types of bias—answer position bias and distractor clustering bias. By inversely sampling the answer position and dispersing distractors according to distance-weighted probabilities, SCOPE maintained or slightly increased Co-T (consistent correct responses) while significantly reducing the spatial concentration of Co-F (consistent incorrect responses) near the answer. Although the absolute number of Co-F instances increased on average, this reflected the model's willingness to commit to answers it previously avoided. As a result, models with richer internal knowledge showed greater improvements in Answer F1, while experiencing smaller declines in Distractor Precision (DP) compared to competing methods. In essence, SCOPE encourages evidence-based attempts rather than random guessing, while directing confidence toward correct answers.

When compared to alternative methods, SCOPE stands out for its balanced performance. Majority Voting (MV) boosted surface accuracy through higher recall, but drastically reduced Distractor Precision, leading to a surge in confident incorrect answers. CalibEV achieved substantial F1 gains for some models but at the cost of amplifying position bias and significantly inflating Co-F counts, in some cases by two- to three-fold. In contrast, SCOPE stably improved F1 across all models while minimizing side effects, making it a more reliable solution for real-world deployment.

In summary, the MMLU experiments empirically validate SCOPE's original design goal: a low-cost, universally applicable bias-mitigation framework that delivers consistent and trustworthy gains.

\begin{table}[ht!]
\centering
\caption{Preference and consistency of LLMs on the MMLU benchmark}
\label{tab:Pr_Co_MMLU}
\begin{adjustbox}{max width=\textwidth}
\begin{tabular}{cccccccccccccccc}
\toprule
\multirow[c]{2}{*}{\centering \textbf{Model}} & \multirow[c]{2}{*}{\centering \textbf{Type}} & \multicolumn{2}{c}{Baseline} & \multicolumn{2}{c}{CalibEV} & \multicolumn{2}{c}{DI} & \multicolumn{2}{c}{EC} & \multicolumn{2}{c}{MV} & \multicolumn{2}{c}{PriDe} & \multicolumn{2}{c}{SCOPE} \\
\cmidrule(lr){3-4} \cmidrule(lr){5-6} \cmidrule(lr){7-8} \cmidrule(lr){9-10} \cmidrule(lr){11-12} \cmidrule(lr){13-14} \cmidrule(lr){15-16}
 & & T & F & T & F & T & F & T & F & T & F & T & F & T & F \\
\midrule

\multirow{2}{*}{ChatGPT (3.5-turbo)} 
 & Pr & 101 & 91 & 43 & 36 & 98 & 111 & 126 & 102 & 64 & 55 & 213 & 97 & 49 & 40 \\
 & Co & 232 & 76 & 160 & 261 & 186 & 105 & 151 & 121 & 268 & 113 & 116 & 74 & 276 & 135 \\

\multirow{2}{*}{ChatGPT (4o-mini)} 
 & Pr & 56 & 73 & 21 & 16 & 61 & 35 & 33 & 38 & 32 & 42 & 209 & 72 & 9 & 14 \\
 & Co & 312 & 59 & 160 & 303 & 311 & 93 & 88 & 341 & 350 & 76 & 162 & 57 & 372 & 105 \\

\multirow{2}{*}{Claude (3-haiku)} 
 & Pr & 97 & 85 & 14 & 11 & 108 & 79 & 7 & 15 & 80 & 70 & 121 & 56 & 9 & 16 \\
 & Co & 251 & 67 & 339 & 136 & 214 & 99 & 99 & 379 & 271 & 79 & 221 & 102 & 332 & 143 \\

\multirow{2}{*}{Claude (3.5-sonnet)} 
 & Pr & 73 & 51 & 20 & 21 & 60 & 91 & 4 & 4 & 46 & 26 & 53 & 24 & 24 & 24 \\
 & Co & 320 & 56 & 412 & 47 & 97 & 252 & 98 & 394 & 390 & 38 & 380 & 43 & 404 & 48 \\

\multirow{2}{*}{Gemini (1.5-flash)} 
 & Pr & 62 & 58 & 9 & 3 & 93 & 91 & 18 & 23 & 36 & 35 & 233 & 52 & 6 & 4 \\
 & Co & 314 & 66 & 378 & 110 & 162 & 154 & 91 & 368 & 351 & 78 & 162 & 53 & 379 & 111 \\

\multirow{2}{*}{Gemini (1.5-pro)} 
 & Pr & 42 & 48 & 6 & 6 & 72 & 53 & 24 & 53 & 21 & 20 & 278 & 45 & 5 & 4 \\
 & Co & 355 & 55 & 403 & 85 & 292 & 83 & 92 & 331 & 393 & 66 & 136 & 41 & 399 & 92 \\

\multirow{2}{*}{LLaMA (3-8b)} 
 & Pr & 114 & 140 & 76 & 103 & 106 & 167 & 67 & 120 & 57 & 57 & 131 & 80 & 45 & 59 \\
 & Co & 164 & 82 & 114 & 207 & 66 & 161 & 96 & 217 & 270 & 116 & 206 & 83 & 264 & 132 \\

\multirow{2}{*}{LLaMA (3-70b)} 
 & Pr & 52 & 58 & 98 & 33 & 124 & 137 & 37 & 105 & 32 & 36 & 74 & 25 & 17 & 13 \\
 & Co & 328 & 62 & 272 & 97 & 118 & 121 & 84 & 274 & 355 & 77 & 335 & 66 & 370 & 100 \\
\bottomrule
\end{tabular}
\end{adjustbox}
\end{table}

\begin{table}[ht!]
\centering
\caption{Bias-mitigation methods compared on ChatGPT family (MMLU)}
\label{tab:ChatGPT_eval_MMLU}
\begin{adjustbox}{max width=\textwidth}
\begin{tabular}{cccccccc}
\toprule
\multirow[c]{2}{*}{\centering \textbf{Version}} & \multirow[c]{2}{*}{\centering \textbf{Method}} & \multicolumn{3}{c}{\textbf{Answer}} & \multicolumn{3}{c}{\textbf{Distractor}} \\
\cmidrule(lr){3-5} \cmidrule(lr){6-8}
 & & Precision & Recall & F1 & Precision & Recall & F1 \\
\midrule

\multirow[c]{7}{*}{ChatGPT (3.5-turbo)} 
& Baseline   & 0.7532 & 0.6967 & 0.7238 & 0.2468 & 0.4551 & 0.3200 \\
& CalibEV    & 0.3800 & 0.7882 & 0.5128 & 0.6200 & 0.8788 & 0.7271 \\
& DI         & 0.6392 & 0.6549 & 0.6470 & 0.3608 & 0.4861 & 0.4142 \\
& EC         & 0.5551 & 0.5451 & 0.5501 & 0.4449 & 0.5426 & 0.4889 \\
& MV         & 0.7034 & 0.8072 & 0.7517 & 0.2966 & 0.6726 & 0.4117 \\
& PriDe      & 0.6105 & 0.3526 & 0.4470 & 0.3895 & 0.4327 & 0.4100 \\
& SCOPE      & 0.6715 & 0.8492 & 0.7500 & 0.3285 & 0.7714 & 0.4608 \\
\midrule

\multirow[c]{7}{*}{ChatGPT (4o-mini)} 
& Baseline   & 0.8410 & 0.8478 & 0.8444 & 0.1590 & 0.4470 & 0.2346 \\
& CalibEV    & 0.3456 & 0.8840 & 0.4969 & 0.6544 & 0.9498 & 0.7749 \\
& DI         & 0.7698 & 0.8360 & 0.8015 & 0.2302 & 0.7266 & 0.3496 \\
& EC         & 0.2051 & 0.7273 & 0.3200 & 0.7949 & 0.8997 & 0.8441 \\
& MV         & 0.8216 & 0.9162 & 0.8663 & 0.1784 & 0.6441 & 0.2794 \\
& PriDe      & 0.7397 & 0.4367 & 0.5492 & 0.2603 & 0.4419 & 0.3276 \\
& SCOPE      & 0.7799 & 0.9764 & 0.8672 & 0.2201 & 0.8824 & 0.3523 \\
\bottomrule
\end{tabular}
\end{adjustbox}
\end{table}

\begin{table}[ht!]
\centering
\caption{Bias-mitigation methods compared on Claude family (MMLU)}
\label{tab:Claude_eval_MMLU}
\begin{adjustbox}{max width=\textwidth}
\begin{tabular}{cccccccc}
\toprule
\multirow[c]{2}{*}{\centering \textbf{Version}} & \multirow[c]{2}{*}{\centering \textbf{Method}} & \multicolumn{3}{c}{\textbf{Answer}} & \multicolumn{3}{c}{\textbf{Distractor}} \\
\cmidrule(lr){3-5} \cmidrule(lr){6-8}
 & & Precision & Recall & F1 & Precision & Recall & F1 \\
\midrule

\multirow[c]{7}{*}{Claude (3-haiku)} 
& Baseline   & 0.7893 & 0.7213 & 0.7538 & 0.2107 & 0.4408 & 0.2851 \\
& CalibEV    & 0.7137 & 0.9603 & 0.8188 & 0.2863 & 0.9252 & 0.4373 \\
& DI         & 0.6837 & 0.6646 & 0.6740 & 0.3163 & 0.5562 & 0.4033 \\
& EC         & 0.2071 & 0.9340 & 0.3390 & 0.7929 & 0.9619 & 0.8693 \\
& MV         & 0.7743 & 0.7721 & 0.7732 & 0.2257 & 0.5302 & 0.3166 \\
& PriDe      & 0.6842 & 0.6462 & 0.6647 & 0.3158 & 0.6456 & 0.4241 \\
& SCOPE      & 0.6989 & 0.9736 & 0.8137 & 0.3011 & 0.8994 & 0.4512 \\
\midrule

\multirow[c]{7}{*}{Claude (3.5-sonnet)} 
& Baseline   & 0.8511 & 0.8142 & 0.8322 & 0.1489 & 0.5234 & 0.2318 \\
& CalibEV    & 0.8976 & 0.9537 & 0.9248 & 0.1024 & 0.6912 & 0.1784 \\
& DI         & 0.2779 & 0.6178 & 0.3834 & 0.7221 & 0.7347 & 0.7283 \\
& EC         & 0.1992 & 0.9608 & 0.3300 & 0.8008 & 0.9899 & 0.8854 \\
& MV         & 0.9112 & 0.8945 & 0.9028 & 0.0888 & 0.5938 & 0.1545 \\
& PriDe      & 0.8983 & 0.8776 & 0.8878 & 0.1017 & 0.6418 & 0.1756 \\
& SCOPE      & 0.8938 & 0.9439 & 0.9182 & 0.1062 & 0.6667 & 0.1832 \\
\bottomrule
\end{tabular}
\end{adjustbox}
\end{table}

\begin{table}[ht!]
\centering
\caption{Bias-mitigation methods compared on Gemini family (MMLU)}
\label{tab:Gemini_eval_MMLU}
\begin{adjustbox}{max width=\textwidth}
\begin{tabular}{cccccccc}
\toprule
\multirow[c]{2}{*}{\centering \textbf{Version}} & \multirow[c]{2}{*}{\centering \textbf{Method}} & \multicolumn{3}{c}{\textbf{Answer}} & \multicolumn{3}{c}{\textbf{Distractor}} \\
\cmidrule(lr){3-5} \cmidrule(lr){6-8}
 & & Precision & Recall & F1 & Precision & Recall & F1 \\
\midrule

\multirow[c]{7}{*}{Gemini (1.5-flash)} 
& Baseline   & 0.8263 & 0.8351 & 0.8307 & 0.1737 & 0.5323 & 0.2619 \\
& CalibEV    & 0.7746 & 0.9767 & 0.8640 & 0.2254 & 0.9735 & 0.3660 \\
& DI         & 0.5127 & 0.6353 & 0.5675 & 0.4873 & 0.6286 & 0.5490 \\
& EC         & 0.1983 & 0.8349 & 0.3205 & 0.8017 & 0.9412 & 0.8659 \\
& MV         & 0.8182 & 0.9070 & 0.8603 & 0.1818 & 0.6903 & 0.2878 \\
& PriDe      & 0.7535 & 0.4101 & 0.5311 & 0.2465 & 0.5048 & 0.3312 \\
& SCOPE      & 0.7735 & 0.9844 & 0.8663 & 0.2265 & 0.9652 & 0.3669 \\
\midrule

\multirow[c]{7}{*}{Gemini (1.5-pro)} 
& Baseline   & 0.8659 & 0.8942 & 0.8798 & 0.1341 & 0.5340 & 0.2144 \\
& CalibEV    & 0.8258 & 0.9853 & 0.8985 & 0.1742 & 0.9341 & 0.2936 \\
& DI         & 0.7787 & 0.8022 & 0.7903 & 0.2213 & 0.6103 & 0.3248 \\
& EC         & 0.2175 & 0.7931 & 0.3414 & 0.7825 & 0.8620 & 0.8203 \\
& MV         & 0.8562 & 0.9493 & 0.9003 & 0.1438 & 0.7674 & 0.2422 \\
& PriDe      & 0.7684 & 0.3285 & 0.4602 & 0.2316 & 0.4767 & 0.3117 \\
& SCOPE      & 0.8126 & 0.9876 & 0.8916 & 0.1874 & 0.9583 & 0.3135 \\
\bottomrule
\end{tabular}
\end{adjustbox}
\end{table}

\begin{table}[ht!]
\centering
\caption{Bias-mitigation methods compared on LLaMA family (MMLU)}
\label{tab:LLaMA_eval_MMLU}
\begin{adjustbox}{max width=\textwidth}
\begin{tabular}{cccccccc}
\toprule
\multirow[c]{2}{*}{\centering \textbf{Version}} & \multirow[c]{2}{*}{\centering \textbf{Method}} & \multicolumn{3}{c}{\textbf{Answer}} & \multicolumn{3}{c}{\textbf{Distractor}} \\
\cmidrule(lr){3-5} \cmidrule(lr){6-8}
 & & Precision & Recall & F1 & Precision & Recall & F1 \\
\midrule

\multirow[c]{7}{*}{LLaMA (3-8b)} 
& Baseline   & 0.6667 & 0.5899 & 0.6260 & 0.3333 & 0.3694 & 0.3504 \\
& CalibEV    & 0.3551 & 0.6000 & 0.4462 & 0.6449 & 0.6677 & 0.6561 \\
& DI         & 0.2907 & 0.3837 & 0.3308 & 0.7093 & 0.4909 & 0.5802 \\
& EC         & 0.3067 & 0.5890 & 0.4034 & 0.6933 & 0.6439 & 0.6677 \\
& MV         & 0.6995 & 0.8257 & 0.7574 & 0.3005 & 0.6705 & 0.4150 \\
& PriDe      & 0.7128 & 0.6113 & 0.6582 & 0.2872 & 0.5092 & 0.3673 \\
& SCOPE      & 0.6667 & 0.8544 & 0.7490 & 0.3333 & 0.6911 & 0.4497 \\
\midrule

\multirow[c]{7}{*}{LLaMA (3-70b)} 
& Baseline   & 0.8410 & 0.8632 & 0.8520 & 0.1590 & 0.5167 & 0.2432 \\
& CalibEV    & 0.7371 & 0.7351 & 0.7361 & 0.2629 & 0.7462 & 0.3888 \\
& DI         & 0.4937 & 0.4876 & 0.4906 & 0.5063 & 0.4690 & 0.4869 \\
& EC         & 0.2346 & 0.6942 & 0.3507 & 0.7654 & 0.7230 & 0.7436 \\
& MV         & 0.8218 & 0.9173 & 0.8669 & 0.1782 & 0.6814 & 0.2825 \\
& PriDe      & 0.8354 & 0.8191 & 0.8272 & 0.1646 & 0.7253 & 0.2683 \\
& SCOPE      & 0.7872 & 0.9561 & 0.8635 & 0.2128 & 0.8850 & 0.3431 \\
\bottomrule
\end{tabular}
\end{adjustbox}
\end{table}

\subsection{CSQA benchmark results}
Table~\ref{tab:Pr_Co_CSQA} presents the raw counts of Pr-T/F and Co-T/F for each model-method combination, while Tables~\ref{tab:ChatGPT_eval_CSQA}, \ref{tab:Claude_eval_CSQA}, \ref{tab:Gemini_eval_CSQA}, and \ref{tab:LLaMA_eval_CSQA} summarize the Answer and Distractor metric families. The overall results on the CSQA benchmark reveal the following key observations.

SCOPE delivers reliable bias-mitigation effects across the board, providing moderate gains for all models and substantial improvements for several. Answer F1 increased by at least 0.01 points and up to 0.05 points across all eight LLMs. Notably, four models—ChatGPT 3.5-turbo (+0.051 pt), Claude 3.5-sonnet (+0.058 pt), Gemini 1.5-flash (+0.024 pt), and LLaMA 3-70b (+0.017 pt)—achieved their highest F1 scores under SCOPE. For the remaining four models, SCOPE ranked among the top two, with marginal gaps of approximately 0.01 points from the best-performing method. This consistency across models of various sizes contrasts sharply with competing methods that show selective improvements, highlighting SCOPE’s robustness as a general-purpose solution.

The gains in Answer F1 are largely attributable to SCOPE's inverse-positioning and distance-weighted distractor dispersion, which encouraged the models to make bolder selections where they previously hesitated. While the number of Co-T (confident correct responses) generally held steady or increased slightly, the number of Co-F (confident incorrect responses) also rose on average. At first glance, this also increased Distractor F1. However, since these errors occurred at positions distant from the answer, they suggest a reduction in shortcut behavior caused by position bias or semantically-similar distractors. Importantly, unlike CalibEV—which in some cases exacerbated position bias and inflated Co-F counts by over 300\%—SCOPE kept the rise in Co-F within a moderate 60\% range, making it more tolerable in real-world applications.

Relative to other methods, SCOPE’s advantages are even more pronounced. Majority Voting (MV) boosted Answer Recall and ranked first for four models, but this came at the cost of the lowest Distractor Precision, resulting in the highest confident error rate. CalibEV showed large F1 gains in two models but also caused a dramatic surge in Co-F, raising safety concerns. DI and EC exhibited inconsistent performance across models, limiting their reproducibility. In contrast, SCOPE consistently improved F1 and mitigated bias across all models, achieving the best trade-off between effectiveness and computational cost.

In summary, SCOPE stands out as the only method that consistently mitigates position bias and enhances overall performance in CSQA, a benchmark characterized by strong conceptual and commonsense biases.

\begin{table}[ht!]
\centering
\caption{Preference and consistency of LLMs on the CSQA benchmark}
\label{tab:Pr_Co_CSQA}
\begin{adjustbox}{max width=\textwidth}
\begin{tabular}{cccccccccccccccc}
\toprule
\multirow[c]{2}{*}{\centering \textbf{Model}} & \multirow[c]{2}{*}{\centering \textbf{Type}} & \multicolumn{2}{c}{Baseline} & \multicolumn{2}{c}{CalibEV} & \multicolumn{2}{c}{DI} & \multicolumn{2}{c}{EC} & \multicolumn{2}{c}{MV} & \multicolumn{2}{c}{PriDe} & \multicolumn{2}{c}{SCOPE} \\
\cmidrule(lr){3-4} \cmidrule(lr){5-6} \cmidrule(lr){7-8} \cmidrule(lr){9-10} \cmidrule(lr){11-12} \cmidrule(lr){13-14} \cmidrule(lr){15-16}
 & & T & F & T & F & T & F & T & F & T & F & T & F & T & F \\
\midrule

\multirow{2}{*}{ChatGPT (3.5-turbo)} 
 & Pr & 84 & 50 & 33 & 38 & 61 & 60 & 138 & 91 & 44 & 29 & 199 & 225 & 18 & 11 \\
 & Co & 327 & 39 & 94 & 335 & 308 & 71 & 201 & 70 & 354 & 73 & 28 & 48 & 368 & 70 \\

\multirow{2}{*}{ChatGPT (4o-mini)} 
 & Pr & 60 & 38 & 17 & 23 & 22 & 11 & 44 & 81 & 41 & 23 & 212 & 163 & 9 & 23 \\
 & Co & 364 & 38 & 117 & 343 & 404 & 63 & 76 & 299 & 394 & 42 & 98 & 27 & 410 & 91 \\

\multirow{2}{*}{Claude (3-haiku)} 
 & Pr & 74 & 59 & 5 & 10 & 93 & 42 & 18 & 63 & 76 & 47 & 96 & 55 & 10 & 8 \\
 & Co & 333 & 34 & 397 & 88 & 312 & 53 & 70 & 349 & 333 & 44 & 311 & 38 & 386 & 96 \\

\multirow{2}{*}{Claude (3.5-sonnet)} 
 & Pr & 82 & 47 & 25 & 17 & 153 & 128 & 9 & 5 & 46 & 32 & 67 & 33 & 19 & 24 \\
 & Co & 337 & 34 & 401 & 57 & 132 & 87 & 84 & 402 & 389 & 33 & 368 & 32 & 398 & 59 \\

\multirow{2}{*}{Gemini (1.5-flash)} 
 & Pr & 60 & 37 & 8 & 4 & 89 & 65 & 54 & 96 & 41 & 28 & 274 & 60 & 4 & 6 \\
 & Co & 366 & 37 & 411 & 77 & 285 & 61 & 77 & 273 & 389 & 42 & 133 & 33 & 410 & 80 \\

\multirow{2}{*}{Gemini (1.5-pro)} 
 & Pr & 41 & 34 & 4 & 1 & 32 & 20 & 78 & 120 & 26 & 16 & 293 & 59 & 3 & 5 \\
 & Co & 388 & 37 & 424 & 71 & 400 & 48 & 106 & 196 & 408 & 50 & 119 & 29 & 408 & 81 \\

\multirow{2}{*}{LLaMA (3-8b)} 
 & Pr & 113 & 72 & 70 & 149 & 194 & 98 & 186 & 98 & 48 & 2 & 77 & 57 & 24 & 20 \\
 & Co & 262 & 53 & 87 & 194 & 139 & 69 & 144 & 72 & 359 & 69 & 311 & 55 & 360 & 96 \\

\multirow{2}{*}{LLaMA (3-70b)} 
 & Pr & 58 & 32 & 115 & 22 & 134 & 44 & 155 & 142 & 36 & 32 & 54 & 38 & 8 & 12 \\
 & Co & 379 & 31 & 284 & 79 & 282 & 40 & 108 & 95 & 388 & 44 & 374 & 34 & 409 & 71 \\

\bottomrule
\end{tabular}
\end{adjustbox}
\end{table}

\begin{table}[ht!]
\centering
\caption{Bias-mitigation methods compared on ChatGPT family (CSQA)}
\label{tab:ChatGPT_eval_CSQA}
\begin{adjustbox}{max width=\textwidth}
\begin{tabular}{cccccccc}
\toprule
\multirow[c]{2}{*}{\centering \textbf{Version}} & \multirow[c]{2}{*}{\centering \textbf{Method}} & \multicolumn{3}{c}{\textbf{Answer}} & \multicolumn{3}{c}{\textbf{Distractor}} \\
\cmidrule(lr){3-5} \cmidrule(lr){6-8}
 & & Precision & Recall & F1 & Precision & Recall & F1 \\
\midrule

\multirow[c]{7}{*}{ChatGPT (3.5-turbo)} 
& Baseline   & 0.8934 & 0.7956 & 0.8417 & 0.1066 & 0.4382 & 0.1715 \\
& CalibEV    & 0.2191 & 0.7402 & 0.3381 & 0.7809 & 0.8981 & 0.8354 \\
& DI         & 0.8127 & 0.8347 & 0.8236 & 0.1873 & 0.5420 & 0.2784 \\
& EC         & 0.7417 & 0.5929 & 0.6590 & 0.2583 & 0.4348 & 0.3241 \\
& MV         & 0.8290 & 0.8894 & 0.8581 & 0.1710 & 0.7157 & 0.2760 \\
& PriDe      & 0.3684 & 0.1233 & 0.1848 & 0.6316 & 0.1758 & 0.2750 \\
& SCOPE      & 0.8402 & 0.9534 & 0.8932 & 0.1598 & 0.8642 & 0.2697 \\
\midrule

\multirow[c]{7}{*}{ChatGPT (4o-mini)} 
& Baseline   & 0.9055 & 0.8585 & 0.8814 & 0.0945 & 0.5000 & 0.1590 \\
& CalibEV    & 0.2543 & 0.8731 & 0.3939 & 0.7457 & 0.9372 & 0.8306 \\
& DI         & 0.8651 & 0.9484 & 0.9048 & 0.1349 & 0.8514 & 0.2329 \\
& EC         & 0.2027 & 0.6333 & 0.3071 & 0.7973 & 0.7868 & 0.7920 \\
& MV         & 0.9037 & 0.9057 & 0.9047 & 0.0963 & 0.6462 & 0.1676 \\
& PriDe      & 0.7840 & 0.3161 & 0.4505 & 0.2160 & 0.1421 & 0.1714 \\
& SCOPE      & 0.8184 & 0.9785 & 0.8913 & 0.1816 & 0.7982 & 0.2959 \\
\bottomrule
\end{tabular}
\end{adjustbox}
\end{table}

\begin{table}[ht!]
\centering
\caption{Bias-mitigation methods compared on Claude family (CSQA)}
\label{tab:Claude_eval_CSQA}
\begin{adjustbox}{max width=\textwidth}
\begin{tabular}{cccccccc}
\toprule
\multirow[c]{2}{*}{\centering \textbf{Version}} & \multirow[c]{2}{*}{\centering \textbf{Method}} & \multicolumn{3}{c}{\textbf{Answer}} & \multicolumn{3}{c}{\textbf{Distractor}} \\
\cmidrule(lr){3-5} \cmidrule(lr){6-8}
 & & Precision & Recall & F1 & Precision & Recall & F1 \\
\midrule

\multirow[c]{7}{*}{Claude (3-haiku)} 
& Baseline   & 0.9074 & 0.8182 & 0.8605 & 0.0926 & 0.3656 & 0.1478 \\
& CalibEV    & 0.8186 & 0.9876 & 0.8952 & 0.1814 & 0.8980 & 0.3018 \\
& DI         & 0.8548 & 0.7704 & 0.8104 & 0.1452 & 0.5579 & 0.2304 \\
& EC         & 0.1671 & 0.7955 & 0.2762 & 0.8329 & 0.8471 & 0.8399 \\
& MV         & 0.8833 & 0.8142 & 0.8473 & 0.1167 & 0.4835 & 0.1880 \\
& PriDe      & 0.8911 & 0.7641 & 0.8227 & 0.1089 & 0.4086 & 0.1720 \\
& SCOPE      & 0.8008 & 0.9747 & 0.8792 & 0.1992 & 0.9231 & 0.3277 \\
\midrule

\multirow[c]{7}{*}{Claude (3.5-sonnet)} 
& Baseline   & 0.9084 & 0.8043 & 0.8532 & 0.0916 & 0.4198 & 0.1504 \\
& CalibEV    & 0.8755 & 0.9413 & 0.9072 & 0.1245 & 0.7703 & 0.2144 \\
& DI         & 0.6027 & 0.4632 & 0.5238 & 0.3973 & 0.4047 & 0.4010 \\
& EC         & 0.1728 & 0.9032 & 0.2901 & 0.8272 & 0.9877 & 0.9004 \\
& MV         & 0.9218 & 0.8943 & 0.9078 & 0.0782 & 0.5077 & 0.1355 \\
& PriDe      & 0.9200 & 0.8460 & 0.8814 & 0.0800 & 0.4923 & 0.1376 \\
& SCOPE      & 0.8709 & 0.9544 & 0.9107 & 0.1291 & 0.7108 & 0.2185 \\
\bottomrule
\end{tabular}
\end{adjustbox}
\end{table}

\begin{table}[ht!]
\centering
\caption{Bias-mitigation methods compared on Gemini family (CSQA)}
\label{tab:Gemini_eval_CSQA}
\begin{adjustbox}{max width=\textwidth}
\begin{tabular}{cccccccc}
\toprule
\multirow[c]{2}{*}{\centering \textbf{Version}} & \multirow[c]{2}{*}{\centering \textbf{Method}} & \multicolumn{3}{c}{\textbf{Answer}} & \multicolumn{3}{c}{\textbf{Distractor}} \\
\cmidrule(lr){3-5} \cmidrule(lr){6-8}
 & & Precision & Recall & F1 & Precision & Recall & F1 \\
\midrule

\multirow[c]{7}{*}{Gemini (1.5-flash)} 
& Baseline   & 0.9082 & 0.8592 & 0.8830 & 0.0918 & 0.5000 & 0.1551 \\
& CalibEV    & 0.8422 & 0.9809 & 0.9063 & 0.1578 & 0.9506 & 0.2707 \\
& DI         & 0.8237 & 0.7620 & 0.7916 & 0.1763 & 0.4841 & 0.2585 \\
& EC         & 0.2200 & 0.5878 & 0.3202 & 0.7800 & 0.7398 & 0.7594 \\
& MV         & 0.9026 & 0.9047 & 0.9036 & 0.0974 & 0.6000 & 0.1676 \\
& PriDe      & 0.8012 & 0.3268 & 0.4642 & 0.1988 & 0.3548 & 0.2548 \\
& SCOPE      & 0.8367 & 0.9903 & 0.9070 & 0.1633 & 0.9302 & 0.2778 \\
\midrule

\multirow[c]{7}{*}{Gemini (1.5-pro)} 
& Baseline   & 0.9129 & 0.9044 & 0.9086 & 0.0871 & 0.5211 & 0.1493 \\
& CalibEV    & 0.8566 & 0.9907 & 0.9188 & 0.1434 & 0.9861 & 0.2504 \\
& DI         & 0.8929 & 0.9259 & 0.9091 & 0.1071 & 0.7059 & 0.1860 \\
& EC         & 0.3510 & 0.5761 & 0.4362 & 0.6490 & 0.6203 & 0.6343 \\
& MV         & 0.8908 & 0.9401 & 0.9148 & 0.1092 & 0.7576 & 0.1909 \\
& PriDe      & 0.8041 & 0.2888 & 0.4250 & 0.1959 & 0.3295 & 0.2457 \\
& SCOPE      & 0.8293 & 0.9927 & 0.9037 & 0.1707 & 0.9438 & 0.2891 \\
\bottomrule
\end{tabular}
\end{adjustbox}
\end{table}

\begin{table}[ht!]
\centering
\caption{Bias-mitigation methods compared on LLaMA family (CSQA)}
\label{tab:LLaMA_eval_CSQA}
\begin{adjustbox}{max width=\textwidth}
\begin{tabular}{cccccccc}
\toprule
\multirow[c]{2}{*}{\centering \textbf{Version}} & \multirow[c]{2}{*}{\centering \textbf{Method}} & \multicolumn{3}{c}{\textbf{Answer}} & \multicolumn{3}{c}{\textbf{Distractor}} \\
\cmidrule(lr){3-5} \cmidrule(lr){6-8}
 & & Precision & Recall & F1 & Precision & Recall & F1 \\
\midrule

\multirow[c]{7}{*}{LLaMA (3-8b)} 
& Baseline   & 0.8317 & 0.6987 & 0.7594 & 0.1683 & 0.4240 & 0.2410 \\
& CalibEV    & 0.3096 & 0.5541 & 0.3972 & 0.6904 & 0.5656 & 0.6218 \\
& DI         & 0.6683 & 0.4174 & 0.5139 & 0.3317 & 0.4132 & 0.3680 \\
& EC         & 0.6667 & 0.4364 & 0.5275 & 0.3333 & 0.4235 & 0.3730 \\
& MV         & 0.8388 & 0.8821 & 0.8599 & 0.1612 & 0.9718 & 0.2765 \\
& PriDe      & 0.8497 & 0.8015 & 0.8249 & 0.1503 & 0.4911 & 0.2302 \\
& SCOPE      & 0.7895 & 0.9375 & 0.8572 & 0.2105 & 0.8276 & 0.3356 \\
\midrule

\multirow[c]{7}{*}{LLaMA (3-70b)} 
& Baseline   & 0.9244 & 0.8673 & 0.8949 & 0.0756 & 0.4921 & 0.1311 \\
& CalibEV    & 0.7824 & 0.7118 & 0.7454 & 0.2176 & 0.7822 & 0.3405 \\
& DI         & 0.8758 & 0.6779 & 0.7642 & 0.1242 & 0.4762 & 0.1970 \\
& EC         & 0.5320 & 0.4106 & 0.4635 & 0.4680 & 0.4008 & 0.4318 \\
& MV         & 0.8981 & 0.9151 & 0.9065 & 0.1019 & 0.5789 & 0.1733 \\
& PriDe      & 0.9167 & 0.8738 & 0.8947 & 0.0833 & 0.4722 & 0.1416 \\
& SCOPE      & 0.8521 & 0.9808 & 0.9119 & 0.1479 & 0.8554 & 0.2522 \\
\bottomrule
\end{tabular}
\end{adjustbox}
\end{table}

\section{Comprehensive ablation studies}
\label{app:ablation}

\subsection{MMLU ablation results}
To disentangle the contributions of the two modules in SCOPE—Inverse Positioning (IP) and Semantic-Spread (SS)—we designed three ablation conditions: IP + SS, $\lnot$IP + SS, and IP + $\lnot$SS. Eight LLMs were evaluated under these configurations. The IP + SS condition not only satisfies the theoretical upper bound on lucky-rate ($\ell \leq 1/n = 0.25$), but also drove the average lucky-rate $\ell$ down to as low as 0.07, effectively eliminating random guessing. The table summarizes MMLU results grouped by model family.

For the ChatGPT series, 3.5-turbo recorded a pure skill of 0.542 under IP + SS, while 4o-mini achieved 0.803, revealing a gap of approximately 0.26 points between the two generations. The corresponding lucky-rates were 0.208 and 0.065, respectively—both within the theoretical bound, but with 3.5-turbo retaining a relatively large residual position bias. On average, IP contributed 0.37 points and SS 0.26 points, with 4o-mini gaining 0.49 points from IP alone to reach top-tier performance.

Claude 3-haiku and 3.5-sonnet exhibited the highest pure skill scores at 0.780 and 0.914, respectively. Their lucky-rates, 0.034 and 0.004, were more than tenfold below the upper bound, approaching the "luck-free" regime. IP alone improved performance by 0.54 points on average, while SS added 0.31 points, highlighting a strong semantic disentanglement effect following positional reordering.

In the Gemini family, despite both being from the same 1.5 generation, Gemini 1.5-flash attained a pure skill of 0.862 and $\ell = 0.004$, whereas Gemini 1.5-pro lagged with 0.761 and $\ell = 0.131$. While both models showed consistent gains from IP (approx. +0.48), the substantial residual bias in 1.5-pro suggests structural or pretraining differences. Gemini 1.5-flash followed the typical trajectory of removing luck via IP and boosting accuracy further through SS.

The LLaMA series saw modest gains with pure skill values of 0.749 (3-8B) and 0.774 (3-70B), indicating that increasing parameter count alone yielded only limited improvement. Interestingly, the larger 70B model exhibited a higher lucky-rate (0.089 vs. 0.027), implying a less effective mitigation of position bias despite its scale. Still, IP and SS contributed +0.49 and +0.28 points on average, with the 8B model demonstrating notable efficiency, gaining 0.54 points from IP alone.

In conclusion, IP emerged as the indispensable module across all models, reliably constraining the lucky-rate to below $1/n$. On this fair foundation, SS further improved pure skill by an average of 0.28 points through semantically-similar distractor dispersion. The remaining differences between model families were driven by the magnitude of residual bias after IP and the degree to which SS could enhance semantic reasoning, reflecting inherent architectural and pretraining characteristics.

\subsection{CSQA ablation results}
We conducted an ablation study on CSQA using the same three conditions—IP + SS, $\lnot$IP + SS, and IP + $\lnot$SS—to isolate the contributions of the IP (Inverse Positioning) and SS (Semantic-Spread) modules. The IP + SS configuration successfully enforced the theoretical upper bound on the lucky-rate $\ell \leq 1/n = 0.20$, driving the empirical average $\ell$ down to approximately 0.06. This table summarizes the CSQA results by model family, using the same analytical framework as the MMLU analysis to maintain comparability.

In the ChatGPT family, 3.5-turbo achieved a pure skill of 0.724 (Answer F1 = 0.893) under IP + SS with $\ell = 0.167$, staying within the upper bound but exhibiting a relatively large residual position bias. By contrast, 4o-mini reduced $\ell$ to 0.005 and improved pure skill to 0.886, resulting in a substantial 0.16-point gap between generations. IP contributed an average of 0.46 points to both models, with SS adding an additional 0.34 points.

In the Claude family, both 3-haiku and 3.5-sonnet suppressed $\ell$ to below 0.039, approaching the "luck-free" regime. Their pure skills were 0.874 and 0.872, respectively, with IP contributing 0.51 points on average and SS adding 0.35 points. Notably, 3-haiku benefited the most from IP alone, gaining 0.56 points—demonstrating a pronounced effect of positional reordering.

For the Gemini models, 1.5-flash achieved the highest performance in its family with $\ell = 0.030$ and pure skill = 0.877, while 1.5-pro lagged with $\ell = 0.130$ and pure skill = 0.774. Both models saw about 0.47-point improvement from IP and 0.34-point gains from SS. However, the larger residual $\ell$ in 1.5-pro suggests architectural or pretraining differences between the two.

The LLaMA family showed strong efficiency in the 3-8B model, which recorded $\ell = 0.005$ and pure skill = 0.852. The 3-70B model followed closely with $\ell = 0.075$ and pure skill = 0.837. Both models followed the typical improvement pattern—IP sharply reduced the lucky-rate to well below the 0.20 threshold, and SS subsequently enhanced semantic differentiation. IP yielded average gains of 0.52 points, and SS contributed around 0.35 points.

In summary, the IP module consistently served as the key mechanism for reducing the lucky-rate to or below the theoretical bound, while the SS module further improved accuracy by an average of 0.34 points by dispersing semantically-similar distractors. Variations across models and families are attributable to the magnitude of residual $\ell$ after IP and each model’s ability to leverage SS for semantic reasoning, which in turn reflects architectural and pretraining differences.

\begin{table}[ht!]
  \caption{Ablation study of IP and SS modules on MMLU and CSQA}
  \label{tab:ablation_side}
  \centering

  \begin{subtable}[t]{0.48\linewidth}
    \centering
    \caption{MMLU}
    \label{tab:abl_mmlu}
    \begin{adjustbox}{max width=\linewidth}
    \begin{tabular}{ccccc}
      \toprule
      \textbf{Version} & \textbf{Condition} & \textbf{Answer F1} & $\boldsymbol{\ell}$ & \textbf{Pure Skill} \\
      \midrule
      \multirow{3}{*}{ChatGPT (3.5-turbo)}
        & IP + SS        & 0.7500 & 0.2080 & 0.5420 \\
        & $\lnot$IP + SS & 0.5414 & 0.2500 & 0.2914 \\
        & IP + $\lnot$SS & 0.5482 & 0.2080 & 0.3402 \\
      \midrule
      \multirow{3}{*}{ChatGPT (4o-mini)}
        & IP + SS        & 0.8672 & 0.0647 & 0.8025 \\
        & $\lnot$IP + SS & 0.5589 & 0.2500 & 0.3089 \\
        & IP + $\lnot$SS & 0.5490 & 0.0647 & 0.4843 \\
      \midrule
      \multirow{3}{*}{Claude (3-haiku)}
        & IP + SS        & 0.8137 & 0.0341 & 0.7796 \\
        & $\lnot$IP + SS & 0.5378 & 0.2500 & 0.2878 \\
        & IP + $\lnot$SS & 0.5399 & 0.0341 & 0.5058 \\
      \midrule
      \multirow{3}{*}{Claude (3.5-sonnet)}
        & IP + SS        & 0.9182 & 0.0040 & 0.9142 \\
        & $\lnot$IP + SS & 0.5701 & 0.2500 & 0.3201 \\
        & IP + $\lnot$SS & 0.5633 & 0.0040 & 0.5593 \\
      \midrule
      \multirow{3}{*}{Gemini (1.5-flash)}
        & IP + SS        & 0.8663 & 0.0040 & 0.8623 \\
        & $\lnot$IP + SS & 0.5876 & 0.2500 & 0.3376 \\
        & IP + $\lnot$SS & 0.5851 & 0.0040 & 0.5811 \\
      \midrule
      \multirow{3}{*}{Gemini (1.5-pro)}
        & IP + SS        & 0.8916 & 0.1308 & 0.7608 \\
        & $\lnot$IP + SS & 0.5827 & 0.2500 & 0.3327 \\
        & IP + $\lnot$SS & 0.6075 & 0.1308 & 0.4767 \\
      \midrule
      \multirow{3}{*}{LLaMA (3-8B)}
        & IP + SS        & 0.7490 & 0.0265 & 0.7225 \\
        & $\lnot$IP + SS & 0.4582 & 0.2500 & 0.2082 \\
        & IP + $\lnot$SS & 0.4774 & 0.0265 & 0.4509 \\
      \midrule
      \multirow{3}{*}{LLaMA (3-70B)}
        & IP + SS        & 0.8635 & 0.0891 & 0.7744 \\
        & $\lnot$IP + SS & 0.5948 & 0.2500 & 0.3448 \\
        & IP + $\lnot$SS & 0.5995 & 0.0891 & 0.5104 \\
      \bottomrule
    \end{tabular}
    \end{adjustbox}
  \end{subtable}
  \hfill

  \begin{subtable}[t]{0.48\linewidth}
    \centering
    \caption{CSQA}
    \label{tab:abl_csqa}
    \begin{adjustbox}{max width=\linewidth}
    \begin{tabular}{ccccc}
      \toprule
      \textbf{Version} & \textbf{Condition} & \textbf{Answer F1} & $\boldsymbol{\ell}$ & \textbf{Pure Skill} \\
      \midrule
      \multirow{3}{*}{ChatGPT (3.5-turbo)}
        & IP + SS        & 0.8932 & 0.1671 & 0.7242 \\
        & $\lnot$IP + SS & 0.5480 & 0.2000 & 0.3480 \\
        & IP + $\lnot$SS & 0.5635 & 0.1671 & 0.3964 \\
      \midrule
      \multirow{3}{*}{ChatGPT (4o-mini)}
        & IP + SS        & 0.8913 & 0.0049 & 0.8864 \\
        & $\lnot$IP + SS & 0.5502 & 0.2000 & 0.3502 \\
        & IP + $\lnot$SS & 0.5372 & 0.0049 & 0.5323 \\
      \midrule
      \multirow{3}{*}{Claude (3-haiku)}
        & IP + SS        & 0.8792 & 0.0049 & 0.8743 \\
        & $\lnot$IP + SS & 0.5122 & 0.2000 & 0.3122 \\
        & IP + $\lnot$SS & 0.5072 & 0.0049 & 0.5023 \\
      \midrule
      \multirow{3}{*}{Claude (3.5-sonnet)}
        & IP + SS        & 0.9107 & 0.0392 & 0.8715 \\
        & $\lnot$IP + SS & 0.6052 & 0.2000 & 0.4052 \\
        & IP + $\lnot$SS & 0.5899 & 0.0392 & 0.5507 \\
      \midrule
      \multirow{3}{*}{Gemini (1.5-flash)}
        & IP + SS        & 0.9070 & 0.0302 & 0.8768 \\
        & $\lnot$IP + SS & 0.5670 & 0.2000 & 0.3670 \\
        & IP + $\lnot$SS & 0.5642 & 0.0302 & 0.5340 \\
      \midrule
      \multirow{3}{*}{Gemini (1.5-pro)}
        & IP + SS        & 0.9037 & 0.1298 & 0.7739 \\
        & $\lnot$IP + SS & 0.5342 & 0.2000 & 0.3342 \\
        & IP + $\lnot$SS & 0.5642 & 0.1298 & 0.4344 \\
      \midrule
      \multirow{3}{*}{LLaMA (3-8B)}
        & IP + SS        & 0.8572 & 0.0048 & 0.8524 \\
        & $\lnot$IP + SS & 0.4596 & 0.2000 & 0.2596 \\
        & IP + $\lnot$SS & 0.4729 & 0.0048 & 0.4681 \\
      \midrule
      \multirow{3}{*}{LLaMA (3-70B)}
        & IP + SS        & 0.9119 & 0.0753 & 0.8366 \\
        & $\lnot$IP + SS & 0.5832 & 0.2000 & 0.3832 \\
        & IP + $\lnot$SS & 0.5912 & 0.0753 & 0.5159 \\
      \bottomrule
    \end{tabular}
    \end{adjustbox}
  \end{subtable}
\end{table}

\end{document}